\documentclass[sigconf]{acmart}

\usepackage{amsmath,amssymb,amsfonts}
\usepackage{graphicx}
\setkeys{Gin}{draft=false} 
\usepackage{multirow}
\usepackage{subcaption}
\usepackage{booktabs}
\usepackage{enumitem}
\usepackage{tikz}
\usetikzlibrary{arrows.meta, positioning, fit, backgrounds, shapes.geometric}

\setcopyright{none}
\acmConference[KDD '27]{Proceedings of the 33rd ACM SIGKDD Conference on Knowledge Discovery and Data Mining}{August 2027}{TBD}
\acmYear{2027}
\acmISBN{}
\acmDOI{}
\begin{document}

\title{Uncertainty Modeling for Multi-Objective RTA Interception with Distillation Acceleration}

\author{Gaoxiang Zhao}
\affiliation{%
  \institution{JD.com}
  \city{Beijing}
  \country{China}
}
\email{zhaogaoxiang.3@jd.com}

\author{Ruinan Qiu}
\affiliation{%
  \institution{JD.com}
  \city{Beijing}
  \country{China}
}
\email{qiuruinan1@jd.com}

\author{Xiaoting Wang}
\authornote{Corresponding author.}
\affiliation{%
  \institution{JD.com}
  \city{Beijing}
  \country{China}
}
\email{wangxiaoting35@jd.com}

\author{Pengpeng Zhao}
\affiliation{%
  \institution{JD.com}
  \city{Beijing}
  \country{China}
}
\email{zhaopengpeng10@jd.com}

\author{Rongjin Wang}
\affiliation{%
  \institution{JD.com}
  \city{Beijing}
  \country{China}
}
\email{wangrongjin3@jd.com}

\author{Zhangang Lin}
\affiliation{%
  \institution{JD.com}
  \city{Beijing}
  \country{China}
}
\email{linzhangang@jd.com}

\author{Xiaoqiang Wang}
\affiliation{%
  \institution{School of Mathematics and Statistics, Shandong University}
  \city{Weihai}
  \country{China}
}
\email{xiaoqiang.wang@sdu.edu.cn}

\renewcommand{\shortauthors}{Zhao et al.}

\begin{abstract}
Real-Time Auction (RTA) interception decides which incoming advertising requests reach downstream systems, and therefore controls the quality of the data those systems learn from. At JD.com, off-site advertising produces on the order of hundreds of billions of requests per day, and the RTA channel alone serves up to hundreds of millions of requests per minute. Filtering low-quality and fraudulent traffic at this scale requires estimating each request's value with calibrated confidence, which we treat as an uncertainty modeling problem. Two obstacles stand in the way. First, advertising labels are severely imbalanced: deals are rare, and we show both analytically and empirically that standard weight-based uncertainty degrades under such sparsity, collapsing onto the predicted probability and adding no signal. Second, methods such as SWAG and Bayesian neural networks require multiple stochastic forward passes per request, making full-traffic scoring prohibitively expensive. We address both problems with UMDA, a multi-objective framework that shares uncertainty across funnel-correlated objectives, using the reliable uncertainty of a balanced metric to compensate for the degenerate uncertainty of sparse ones. We then distill the multi-pass teacher into a single-pass student that reproduces both aleatoric and epistemic uncertainty at roughly one-tenth of the inference cost. On JD e-commerce dataset and the public Criteo dataset, UMDA supplies more effective samples to downstream tasks, and the distilled student preserves this capability. In production, it scores the full traffic in a near-line pipeline that feeds an hourly blacklist for online interception; a seven-day A/B test on 5\% of live traffic cuts the click fraud rate by 3.59\% and raises CVR by 4.01\% at a matched interception ratio while leaving converted users essentially unchanged, and the model has since been deployed to full traffic.
\end{abstract}



\maketitle

\section{Introduction}
In online advertising, RTA mechanisms determine which traffic is exposed to downstream systems. The scale is substantial: off-site advertising in JD.com's system generates on the order of hundreds of billions of requests per day, and the RTA channel alone serves up to hundreds of millions of requests per minute. At this scale, an interception model must meet strict throughput and cost constraints as well as be accurate. Since not all traffic contributes equally to campaign performance, interception should filter out unproductive requests while preserving those that match the advertiser's objectives. This is difficult: it requires accurate prediction of several user-behavior metrics together with dependable estimates of prediction confidence under highly dynamic conditions. A natural approach is to combine multi-objective optimization with uncertainty modeling. Multi-objective methods such as PLE \cite{PLE}, MMOE \cite{MMOE}, and ESMM \cite{ESMM} improve performance by sharing knowledge across related tasks such as Click-Through Rate (CTR) and Conversion Rate (CVR) prediction. Recent extensions further consider multi-scenario modeling \cite{M2M} or explicit action-topic disentanglement \cite{MVKE}. Uncertainty modeling techniques such as Bayesian Neural Networks (BNN) \cite{BNN}, Monte Carlo Dropout \cite{MC_Dropout}, and SWAG \cite{SWAG} quantify prediction confidence by treating network weights as distributions and estimating variance through repeated inference. Combining the two lets RTA interception rank requests by their predicted contribution to downstream objectives while using uncertainty to avoid discarding valuable but ambiguous traffic, improving both the effectiveness and the reliability of traffic selection.

In RTA interception, multi-objective modeling jointly optimizes several objectives for accurate predictions, while uncertainty modeling identifies high-uncertainty traffic, allowing it to pass through to downstream systems to avoid losing potentially valuable customers. A key challenge for uncertainty modeling is label imbalance \cite{Imbalance_labels}, which causes biased pass-through. BNNs and SWAG capture uncertainty effectively but require multiple passes, creating a computational bottleneck. EDL \cite{EDL} generates probabilities and uncertainty in one pass, yet primarily captures epistemic uncertainty and struggles with aleatoric uncertainty, yielding inferior estimates. DDU \cite{DDU} offers an alternative one-pass method that captures both uncertainties, but requires residual connections and spectral normalization, constraining architecture.

To overcome these limitations, we target two central challenges: the biased traffic passing caused by label imbalance, and the computational burden of repeated forward passes in standard uncertainty modeling. For the first, we introduce a method that leverages uncertainty estimates from balanced metrics to guide the traffic passing for imbalanced ones, within a multi-objective optimization framework that adaptively manages both balanced and imbalanced objectives and shares uncertainty across the metrics. For the second, especially critical under the throughput and cost constraints of full-traffic scoring, we draw on knowledge distillation \cite{Hinton2015} and recent multi-task distillation methods \cite{DMMP, CrossDistil} to distill the teacher's multi-pass uncertainty into a single-pass student. Specifically, we propose a distilled model that jointly learns to predict the target labels and the teacher model's uncertainty outputs, enabling efficient single-pass uncertainty estimation. Our main contributions are summarized as follows:

\begin{itemize}[leftmargin=1.2em, topsep=2pt, itemsep=2pt, parsep=0pt]
	\item \textbf{Analysis of why uncertainty fails under sparsity.} We analyze the internal mechanism of weight-based uncertainty modeling and show, both analytically and empirically, that the uncertainty predicted for each sample is shaped by both the sample's properties and the conditional probabilities of its neighboring samples; under extreme label sparsity it collapses onto the predicted probability and ceases to carry additional signal.
	\item \textbf{UMDA, an uncertainty-sharing framework.} We propose a unified framework that combines multi-objective optimization with uncertainty modeling and shares uncertainty across funnel-correlated objectives, using the trustworthy uncertainty of a balanced metric to compensate for the degenerate uncertainty of sparse ones, thereby estimating usable uncertainty under both balanced and imbalanced labels.
	\item \textbf{Distillation as a deployment enabler.} We introduce a distilled uncertainty model that reproduces both aleatoric and epistemic uncertainty in a single forward pass, retaining traffic-passing capability while achieving a 10x inference speedup. Distillation is not merely an optimization but a prerequisite for deployment: the original UMDA requires roughly eleven stochastic forward passes per request, making full-traffic scoring economically infeasible; the distilled model is what allows the entire traffic to be scored within the production compute budget.
	\item \textbf{Deployment at industrial scale.} We deploy the distilled model in JD.com's production advertising system, which serves up to hundreds of millions of RTA requests per minute, and report a seven-day online A/B test on 5\% of live traffic---after which the model was rolled out to full production traffic---together with the system architecture, design tradeoffs, and lessons learned from running uncertainty-aware interception in production.
\end{itemize}

\section{Related Work}
Related works can be broadly grouped into three categories: (i) uncertainty modeling, (ii) multi-objective prediction, and (iii) approaches that integrate both uncertainty and multi-objective learning to simultaneously improve predictive performance and estimate model confidence.

\subsection{Uncertainty modeling}
Uncertainty modeling quantifies a model's confidence in its predictions; approaches relying solely on cross-entropy probabilities often lead to overconfidence, whereas uncertainty modeling provides more reliable estimates and improves generalization. A recent critical survey \cite{uncertainty_survey} categorizes different uncertainty types and their integration into AI systems. Some methods directly predict uncertainty by assuming a Gaussian output $y \sim \mathcal{N}(\mu(x), \sigma^2(x))$ trained with $\mathcal{L}(x, y) = \frac{1}{2} \log \sigma^2(x) + (y - \mu(x))^2 / (2\sigma^2(x))$; while efficient, they rely on strong distributional assumptions and capture only aleatoric uncertainty. As noted in the Introduction, DDU instead produces both aleatoric and epistemic uncertainties in a single pass, but at the cost of architectural constraints from its spectral-normalized residual feature extractor.

Another class treats each weight as a distribution: sampling repeatedly, multiple forward passes yield outputs whose variance quantifies uncertainty. Bayesian Neural Networks (BNNs) assign a variational distribution $q_{\theta}(w)$ and infer the posterior $p(w \mid D) \propto p(D \mid w)\,p(w)$ by minimizing the KL divergence; the trade-offs between fidelity and efficiency in such approximate inference are surveyed in \cite{uncertainty_survey1}. As cheaper alternatives, MC Dropout implicitly samples from a weight distribution by randomly dropping neurons at inference, while SWAG constructs a weight distribution from training trajectories.

\subsection{Multi-Objective modeling}
Multi-objective learning in recommendation has evolved from dependency modeling to dynamic frameworks. ESMM \cite{ESMM} resolves CVR sparsity by decomposing it into CTR and post-click conversion; MMoE \cite{MMOE} employs shared experts with task-specific gating to reduce negative transfer, while PLE \cite{PLE} introduces progressive layered extraction to disentangle shared/task-specific experts. Subsequent work explicitly models task dependencies: AITM \cite{AITM} adaptively transfers knowledge from CTR to CVR, SNR \cite{SNR} dynamically allocates representations via a neural router, and GRec \cite{GRec} enhances scalability with larger expert pools and sparse gating. MetaBalance \cite{MetaBalance} uses meta-learning for dynamic task weighting, and AutoMTL \cite{AutoMTL} unifies NAS, transfer learning, and dynamic routing for systematic multi-task decomposition.

\subsection{Joint modeling of multi-objective learning and uncertainty quantification}
Recent work combines multi-objective learning with uncertainty quantification for robust decisions. Task-balancing methods such as GradNorm \cite{GradNorm} and DWA \cite{DWA} dynamically reweight objectives to prevent dominant tasks from dictating the optimization, while Kendall et al.\ \cite{Kendall2018} use learnable uncertainty parameters to balance noisy objectives, and UKD \cite{UKD} extends this to semi-supervised ad learning: a teacher generates pseudo-labels for unclicked ads, and a student uses predictor discrepancy to down-weight high-uncertainty samples. Both show that jointly estimating targets and confidence improves accuracy and reliability.

A large body of deployed industrial systems has shaped how large-scale e-commerce and advertising pipelines retrieve, rank, and filter traffic---from billion-scale commodity embeddings \cite{EGES} and embedding-based retrieval in Taobao \cite{TaobaoEBR} and Facebook search \cite{FacebookEBR}, to unified retrieval frameworks such as UniERF \cite{UniERF} and hard-won lessons from production click-prediction systems \cite{Trenches}. These works establish the engineering patterns---candidate generation, multi-channel retrieval, and cost-aware serving---within which RTA interception operates, but they optimize retrieval or ranking quality rather than the confidence-aware traffic filtering we target.

Our work differs from these lines in three respects, motivated by the constraints of industrial RTA interception. First, rather than treating uncertainty as a per-task regularizer, we \emph{share} uncertainty across funnel-correlated objectives, using the well-estimated uncertainty of a balanced metric to compensate for the unreliable uncertainty of sparse ones. Second, we provide an analytical account of \emph{why} weight-based uncertainty degrades under label imbalance, rather than treating it as a black box. Third, we address the serving cost that these methods leave open: we distill multi-pass uncertainty into a single forward pass, making it feasible to score the full traffic within the production compute budget.

\section{Method}
In this section, we first formulate the interception scenario and motivate uncertainty modeling. We then analyze uncertainty estimation in advertising auctions, showing its robustness over probabilistic outputs and its dependence on data balance, and characterizing when it succeeds and fails. This analysis is what motivates the architectural choices that follow: on its basis, we integrate multi-objective modeling with uncertainty to overcome the imbalanced-label limitation and present the overall UMDA architecture. Finally, we describe the distillation framework that improves efficiency while preserving uncertainty measurement, which is the component we ultimately deploy in production. Throughout, we keep the exposition close to the constraints of the production setting, since the same throughput and cost limits that motivate distillation also shape the earlier modeling choices.

\subsection{Interception scenario formalization}
We first present the general problem of the RTA interception scenario. The interception scenario aims to maximize the quality of downstream traffic while limiting the interception rate. In this context, the optimization goal is to selectively block low-quality traffic to enhance overall traffic quality:
\begin{equation}
	\begin{aligned}
		\text{obj.} \quad & \max \sum_i v_i \cdot z_i \\
		\text{s.t.} \quad & \frac{\sum_i (1 - z_i)}{N} = r \\
		& z_i \in \{0,1\},
	\end{aligned}
\end{equation}
where $i \in \{1,\dots,N\}$ and $N$ is the total number of samples. Let $z_i \in \{0,1\}$ denote the intercept/pass decision for the $i$-th sample, with $z_i = 1$ indicating a pass. Let $r$ represent our interception ratio, which varies dynamically with the advertising budget and consumption. Due to the sufficient traffic, we select the best $(1 - r)$ fraction of traffic to pass downstream in order to enhance conversion efficiency. The expected reward of the $i$-th sample, denoted by $v_i$, is jointly evaluated based on four metrics: C2S click, online, add to cart, and deal. Let $t_{i,1}, t_{i,2}, t_{i,3}, t_{i,4}$ be the scores of the $i$-th sample on these four metrics, respectively. Then, we have $v_i = f(t_{i,1}, t_{i,2}, t_{i,3}, t_{i,4})$.

As an aggregation of the four individual scores, the function $f$ can be flexibly adjusted according to the specific emphasis required in different application scenarios. The metrics C2S (user click on external ads), online (landing on the e-commerce platform post-click), add to cart, and deal form a hierarchical chain of user actions. Within a multi-objective modeling framework, each metric is equipped with its own uncertainty estimate. For the $k$-th metric of the $i$-th sample, we define a combined score $t_{i,k} = h(s_{i,k}, u_{i,k})$, where $s_{i,k}$ is the predicted probability for the $k$-th metric of the $i$-th sample, $u_{i,k}$ the associated uncertainty, and $h$ is designed to increase with both $s_{i,k}$ and $u_{i,k}$, thereby encouraging higher prediction scores while ensuring that outputs with high uncertainty are not intercepted. This design allows the system to prioritize intercepting low-quality traffic—defined as traffic with low predicted probability and high confidence—while reserving traffic with low predicted probability but high uncertainty, as it may hold significant value. In our specific scenario, $h$ is implemented as a two-stage policy: we first route the portion of traffic with the highest uncertainty downstream for further evaluation; among the remaining traffic, we then allow those samples with the highest predicted probabilities to pass.

\subsection{Approximate estimation of uncertainty outputs}
\label{sec:approx}
Uncertainty modeling is regarded as capable of quantifying the uncertainty of a model's output, yet it typically performs poorly under data imbalance. To elucidate this and guide model design, we approximate how uncertainty estimation relates to the gradients in the vicinity of sample points, and find that under imbalance the output of uncertainty modeling is highly coupled with the predicted probability, thereby failing to provide additional information.

We focus on ambiguous samples—those that share exactly the same features yet exhibit inconsistent labels. This phenomenon arises because categorical variables typically dominate in advertising scenarios, and the same user often engages in multiple distinct visit behaviors whose labels are generally not consistent. Formally, the conditional probability is defined as
\begin{equation}
	q = P(y=1 \mid x^*) = \frac{1}{m} \sum_{i=1}^{m} y_i, \quad 0 < q < 1,
	\label{eq:condprob}
\end{equation}
where $y$ is the true label. The conditional probability measures how likely the true label equals 1 given the occurrence of the features; for ambiguous samples it lies strictly between 0 and 1.

Analyzing how the prediction error evolves under stochastic gradient descent (SGD), and modeling its dynamics as an AR(1) process, we obtain a closed-form expression for the predictive uncertainty (the full derivation is provided in Appendix~\ref{app:derivation}):
\begin{equation}\label{uncertainty_expression1}
	\mathrm{Var}(e_t) = \frac{\eta c}{2 - \eta c\, q (1-q)},
\end{equation}
where $\eta$ is the learning rate and $c$ is the (squared) local gradient norm at the sample. This result reveals that the predictive uncertainty is determined by two factors: the term $q(1-q)$, which peaks at $q=0.5$ and symmetrically decays towards $0$ and $1$, and the local gradient norm $c$, which depends on the sample's features and its neighborhood. To characterize $c$, we further analyze how a neighboring sample $x_2$ (with true conditional probability $p_2$) affects the local gradient at $x_1$. Using a Hessian-based estimate (Appendix~\ref{app:derivation}), the relative change in the gradient norm—and hence the uncertainty—is proportional to
\begin{equation}\label{neighbor1}
	(1 - 2q_1)(p_2 - q_1)\, \phi_1^\top \phi_2,
\end{equation}
where $q_1$ is the predicted probability at $x_1$ and $\phi_1,\phi_2$ are the feature-layer representations of the two samples. Equation~\eqref{neighbor1} reveals two key insights: (1) the influence scales with feature similarity $\phi_1^\top \phi_2$; (2) the direction of change depends on the mismatch between $q_1$ and $p_2$: if $q_1 < 0.5$ (leaning toward the negative class), uncertainty increases when the neighbor leans more positive ($p_2 > q_1$) and decreases when the neighbor leans more negative ($p_2 < q_1$). This formally underpins the intuition that uncertainty modeling is not merely a pointwise property but inherently reflects the label consistency of the surrounding training data.

\subsection{Analysis of effective and ineffective uncertainty scenarios}
By analyzing \eqref{uncertainty_expression1} and \eqref{neighbor1}, the role of uncertainty modeling can be decomposed into two parts. The first part is the predicted probability, where uncertainty increases as the predicted probability approaches 0.5. The second part is the influence of neighboring samples on $c$, which can be regarded as a correction strategy $(1 - 2q_1)(p_2 - q_1)\, \phi_1^\top \phi_2$.

Under severe class imbalance where the positive class is rare, a key failure of uncertainty modeling arises from its reliance on the baseline $q(1-q)$. This formulation implicitly uses 0.5 as the reference point for maximum uncertainty. However, model predictions in such settings converge toward the empirical base rate $\pi = N_+/N$, where $N_+$ is the number of positive samples, which satisfies $\pi \ll 0.5$. Consequently, $q \approx \pi$ rather than 0.5. This discrepancy causes samples whose predicted probability $q$ is near 0.5 (i.e., $q \gg \pi$) to be incorrectly flagged as high-uncertainty—even though a prediction of $q \approx 0.5$ already represents a strong indication of the positive class. This misclassification misleads downstream decision-making. Furthermore, the correction term $c$ fails in sparse regimes: when the indicator is sparse, $(1-2q_1) > 0$ holds, and if the neighboring sample $p_2$ indicates the positive class more strongly than $q_1$, the uncertainty estimate paradoxically increases, rendering the correction term ineffective.

\subsection{UMDA}
\begin{figure*}[t]
	\centering
	\includegraphics[width=0.85\textwidth]{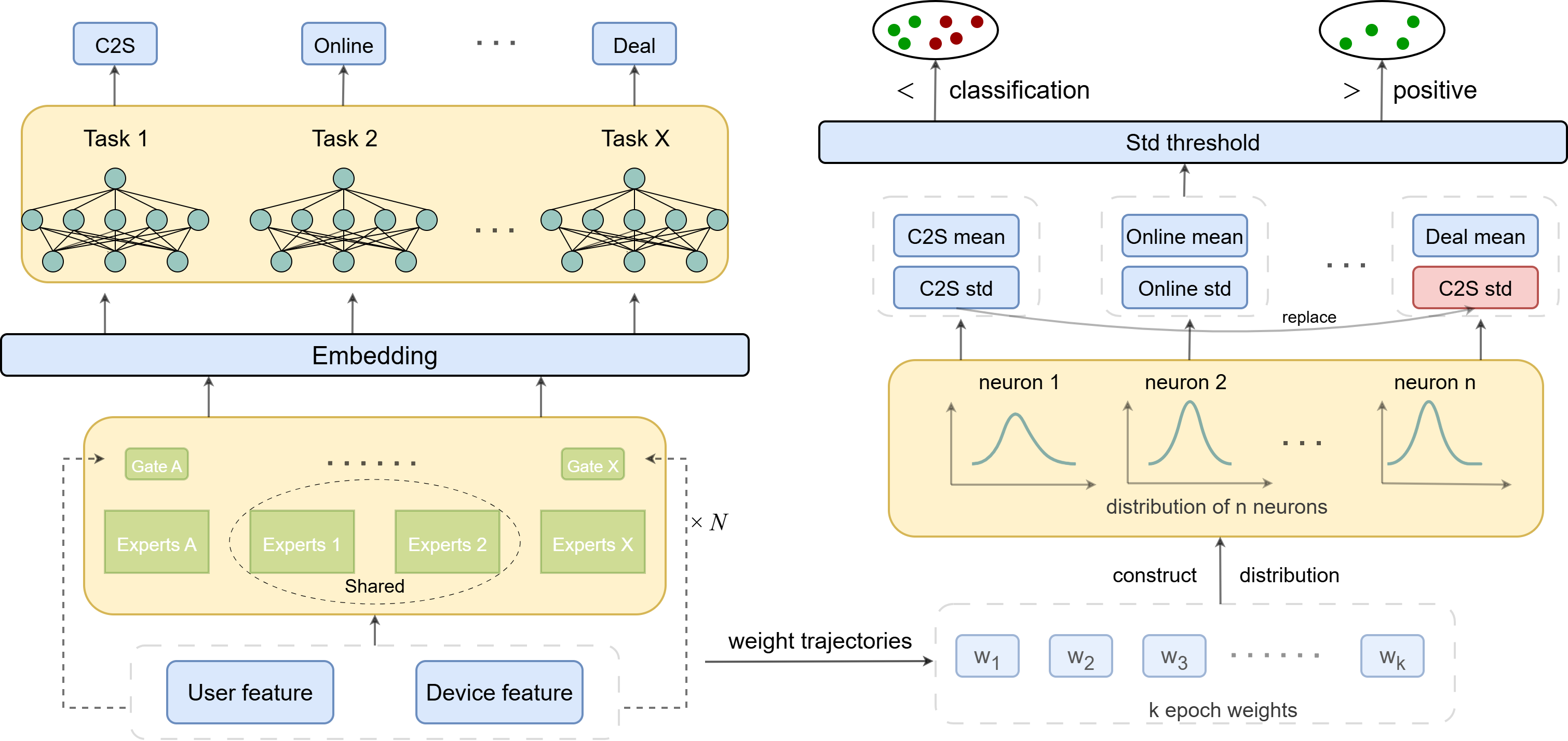}
	\caption{The main network of the UMDA model (with PLE and SWAG as examples). PLE models the output probabilities of multiple distinct metrics, while SWAG establishes uncertainty for each metric and aims to substitute the uncertainty estimates of imbalanced metrics with those of balanced metrics that exhibit consistent uncertainty.}
	\label{fig:overall_performance}
\end{figure*}

To address the failure of uncertainty modeling on imbalanced samples, the most direct approach is to replace it with a feasible uncertainty estimate. We construct a straightforward alternative that leverages the funnel nature of tasks in advertising scenarios, combines multi-objective modeling with uncertainty modeling, outputs uncertainties for multiple metrics, and allows them to be shared, thereby resolving this issue. The uncertainty modeling approach can be extended to various methods such as Deep Ensembles, SWAG, and MC Dropout, and the multi-objective co-learning method is also arbitrary. To present the structural design more directly, we adopt SWAG and PLE as illustrative architectures for uncertainty modeling and multi-objective modeling, respectively, and provide the main design diagram of UMDA in Figure~\ref{fig:overall_performance}.

UMDA is a two-stage model designed to integrate multi-objective learning with uncertainty modeling for interception scenarios. In the first stage, a multi-objective model is trained across multiple targets to capture task-specific patterns and dependencies. The model follows a Progressive Layered Extraction (PLE) architecture, where each task's representation is computed through layer-wise transformations. For the $t$-th target at layer $l$, the propagation is defined as $h^{(l, t)} = f^{(l, t)}(h^{(l-1, t)}, g_{\text{shared}}^{(l-1)})$, where $h^{(l, t)}$ is the task-specific hidden state for target $t$ at layer $l$, $g_{\mathrm{shared}}^{(l-1)}$ is the shared representation from the previous layer, and $f^{(l, t)}$ denotes the task-specific transformation at layer $l$. The initial hidden state $h^{(0, t)}$ is derived from the input features.

In the second stage, uncertainty is estimated using the weight trajectories collected during training via SWAG. We sample weights $w_s$ from the approximate posterior as
\begin{equation}
	w_s = \mu_{\text{SWA}} + \tfrac{1}{\sqrt{2}} \Sigma_{\text{diag}}^{1/2} \zeta_1^{(s)} + \tfrac{1}{\sqrt{2(R-1)}} D \zeta_2^{(s)},
\end{equation}
where $\mu_{\text{SWA}}$ is the stochastic weight average, $\Sigma_{\text{diag}}$ is the diagonal covariance estimate, $D$ contains the top $R$ deviation vectors from the SWA trajectory, and $\zeta_1^{(s)}, \zeta_2^{(s)} \sim \mathcal{N}(0, I)$ are standard Gaussian noise vectors for the $s$-th sample. Multiple samples ($s = 1, \dots, S$) are drawn to estimate predictive uncertainty.

UMDA allows uncertainty to be modeled across multiple targets simultaneously; note that in our scenario, the uncertainty of the deal label is replaced by the uncertainty of the C2S click label. The rationale is twofold. First, the four metrics form a hierarchical funnel (C2S click $\rightarrow$ online $\rightarrow$ deal): a deal is necessarily preceded by a click, so the two labels are structurally coupled and a sample whose click behavior is ambiguous tends to be ambiguous at the deal stage as well. Second, Section~\ref{sec:approx} shows that weight-based uncertainty is reliable only when predictions avoid base-rate collapse—true for the balanced click metric, but not for the sparse deal metric, whose uncertainty (Figure~\ref{fig:uncertainty_all}) merely tracks its confidence. Replacing the sparse metric's degenerate uncertainty with the balanced metric's sound estimate therefore passes a useful signal down the funnel without manufacturing one. It should be noted that UMDA is not restricted to a single method for combining multi-objective learning and uncertainty. Provided that the uncertainty method constructs or approximates a weight distribution and that the metrics are correlated, UMDA can be extended to the combinations of methods described above. This is further validated in the experiments.

\subsection{Distillation of UMDA}
\label{sec:distill}
UMDA provides reliable uncertainty estimation and strong interception capability. However, uncertainty modeling requires multiple inferences during prediction, which makes scoring the full traffic prohibitively expensive. To address this issue, we propose a dual-objective submodel for each target that jointly learns the true classification labels and the uncertainty outputs provided by UMDA. Prior work (DDU) has shown that a single forward pass can capture both aleatoric and epistemic uncertainty, confirming that learning uncertainty through a single-pass distilled model is feasible and reliable.

\begin{figure*}[t]
	\centering
	\includegraphics[width=0.85\textwidth]{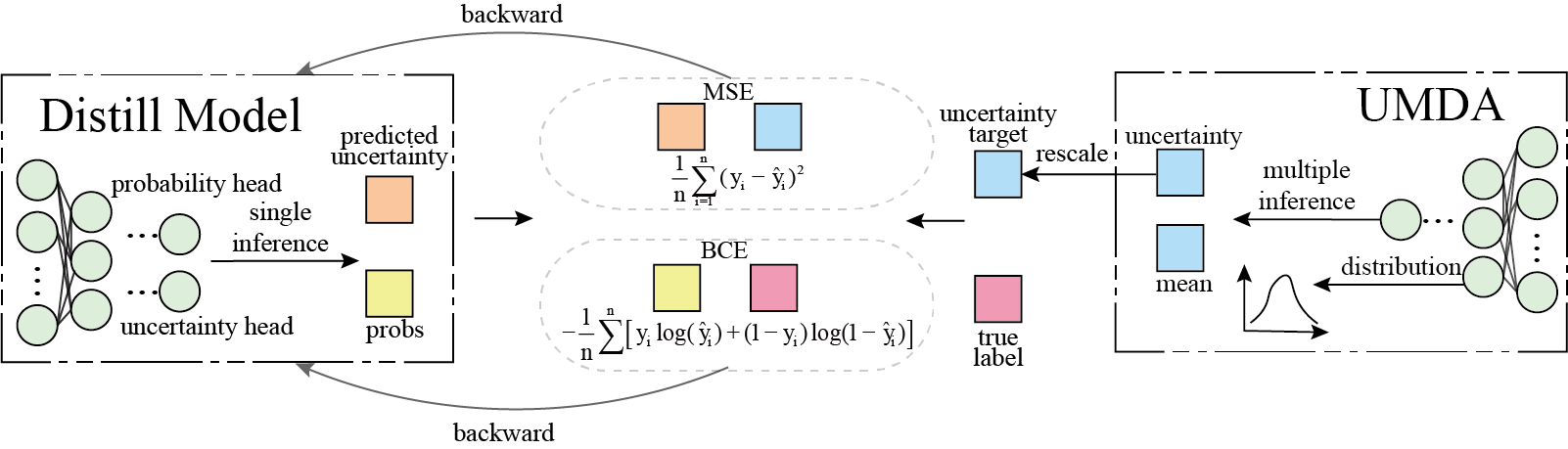}
	\caption{Overview of the proposed uncertainty distillation framework. The predictive variance from the original UMDA model serves as the uncertainty target, which is rescaled and combined with the classification labels to form dual learning objectives for the distilled submodel.}
	\label{fig:distill_model}
\end{figure*}

As illustrated in Figure~\ref{fig:distill_model}, the distillation framework proceeds as follows. The original UMDA model performs multiple stochastic forward passes to obtain the predictive mean and variance; the variance serves as the uncertainty estimate for each submodel. This uncertainty, together with the ground-truth labels, forms the dual learning targets for the distilled model. Since the distilled uncertainty labels and classification labels typically differ in magnitude, directly training on both objectives would cause the classification loss to dominate, leaving insufficient signal for uncertainty learning. Given that uncertainty-based interception relies on relative proportions rather than absolute values, we propose a heuristic label adjustment strategy. Let $y_i \in \{0,1\}$ denote the classification label of sample $i$, and $u_i$ the corresponding output uncertainty. We first compute the mean label $\bar{y} = \frac{1}{N}\sum_{i=1}^N y_i$ and the mean uncertainty $\bar{u} = \frac{1}{N}\sum_{i=1}^N u_i$ over all $N$ samples. The ratio of these two means is $\rho = \bar{y} / \bar{u}$, and every uncertainty value is rescaled by this ratio:
\begin{equation}
	\tilde{u}_i = \rho \cdot u_i \quad \text{for all } i = 1,\dots,N.
\end{equation}
This rescaling enforces $\frac{1}{N}\sum_{i=1}^N \tilde{u}_i = \bar{y}$, aligning the average uncertainty with the mean label. Under this adjustment, the model can effectively learn both the predicted probabilities and the uncertainty estimates simultaneously.

\section{Experiments}
\label{sec:exp}
We first validate two theoretical properties: (i) the uncertainty estimate is centered around a prediction probability of 0.5, and (ii) it is influenced by the density of surrounding samples. Under extreme label sparsity, the centering around 0.5 collapses onto the predicted probability, causing a failure mode. We then show that UMDA alleviates this via uncertainty sharing, providing more reliable estimates for sparse instances and benefiting downstream tasks through indirect passing. Finally, we evaluate the distilled model and validate its practical value via an online A/B test (Section~\ref{sec:deployment}), confirming that it preserves downstream capability while substantially cutting inference cost.

\subsection{Datasets}
The primary dataset is collected from JD advertising system. Each traffic instance contains 108 features covering both user and device-level attributes. From this pool, we randomly sample 1 million instances. To contextualize the sparsity of this dataset, we note the following distribution among the four behavioral metrics: the C2S click is the most balanced, whereas the online metric is relatively more imbalanced compared to the C2S click. Both the add-to-cart and deal metrics are extremely sparse, with deal being the rarest positive event. The second dataset, subsampled from Criteo \cite{criteo1tb} with click rate fixed at ~50\%, contains 1,153,684 samples. Each sample has 12 features and three labels—click, conversion (requiring a prior click), and exposure—with post-sampling conversion and exposure rates of 3.18\% and 16.50\%, respectively. Both datasets use an 8:2 train-test split.

\subsection{Verification of uncertainty properties}
We begin by empirically validating two theoretical properties established in our framework derivation using the Criteo Click dataset. The first property states that the uncertainty estimate naturally centers around a predicted probability of 0.5, reflecting maximum epistemic ambiguity at the decision boundary. The second property reveals that under extreme sample sparsity, uncertainty quantification becomes fundamentally compromised: the model erroneously assigns high uncertainty to samples that otherwise receive high predicted probabilities. We illustrate the relationship between predicted probability and its uncertainty estimate for each of the four behavioral metrics in Appendix~\ref{app:uncertainty_props}, Figure~\ref{fig:uncertainty_all}.

The results in Figure~\ref{fig:uncertainty_all} provide empirical confirmation of the two theoretical properties. For the two relatively balanced metrics, C2S click and Online, the uncertainty estimates exhibit the expected behavior: uncertainty peaks at a predicted probability of approximately 0.5 and diminishes symmetrically toward both extremes. This aligns precisely with the theoretical property that epistemic uncertainty should maximize near the decision boundary.

In stark contrast, the two extremely sparse metrics, Add to cart and Deal, display a markedly different relationship, where uncertainty increases monotonically with predicted probability. Quantitatively, the Pearson correlation coefficients are 0.887 for Add to cart and 0.932 for Deal. This strong linear dependency indicates that, under extreme sparsity, high-uncertainty samples substantially overlap with high-confidence ones. Consequently, uncertainty quantification loses its distinct informativeness—rather than identifying a unique subset of useful instances, it merely recapitulates the model's confidence scores, adding little value for sample selection.

To further elucidate the origins of these two distinct patterns, we note that if uncertainty were determined solely by the conditional probability, it should exhibit a strict inverted-V symmetry centered at 0.5. The empirical behaviors across the four metrics fail to display this idealized shape, indicating that uncertainty estimation is not merely a function of pointwise conditional probabilities but is significantly modulated by the local geometric structure of the sample space.

Such dependency naturally arises in uncertainty measures that rely on variations in model weights, since the predictive variance is influenced by the density of neighboring instances and their labels. To isolate this effect, we conducted a controlled analysis on the C2S click metric: among test samples with conditional probability exactly 0.5, we examined their neighbors within a Euclidean distance of 1 and found that the estimated uncertainty of the central samples shifts systematically as the mean conditional probability of these neighbors deviates from 0.5 in either direction (Appendix~\ref{app:neighbor}). This confirms that uncertainty values are not mere reiterations of point estimates but are profoundly influenced by the local consistency of the training data: when the surrounding environment overwhelmingly favors a specific class, the model's uncertainty near the decision boundary is correspondingly suppressed.

\subsection{Evaluation of passing capability}
\begin{figure}[t]
	\centering
	\begin{subfigure}[b]{0.49\columnwidth}
		\centering
		\includegraphics[width=\textwidth]{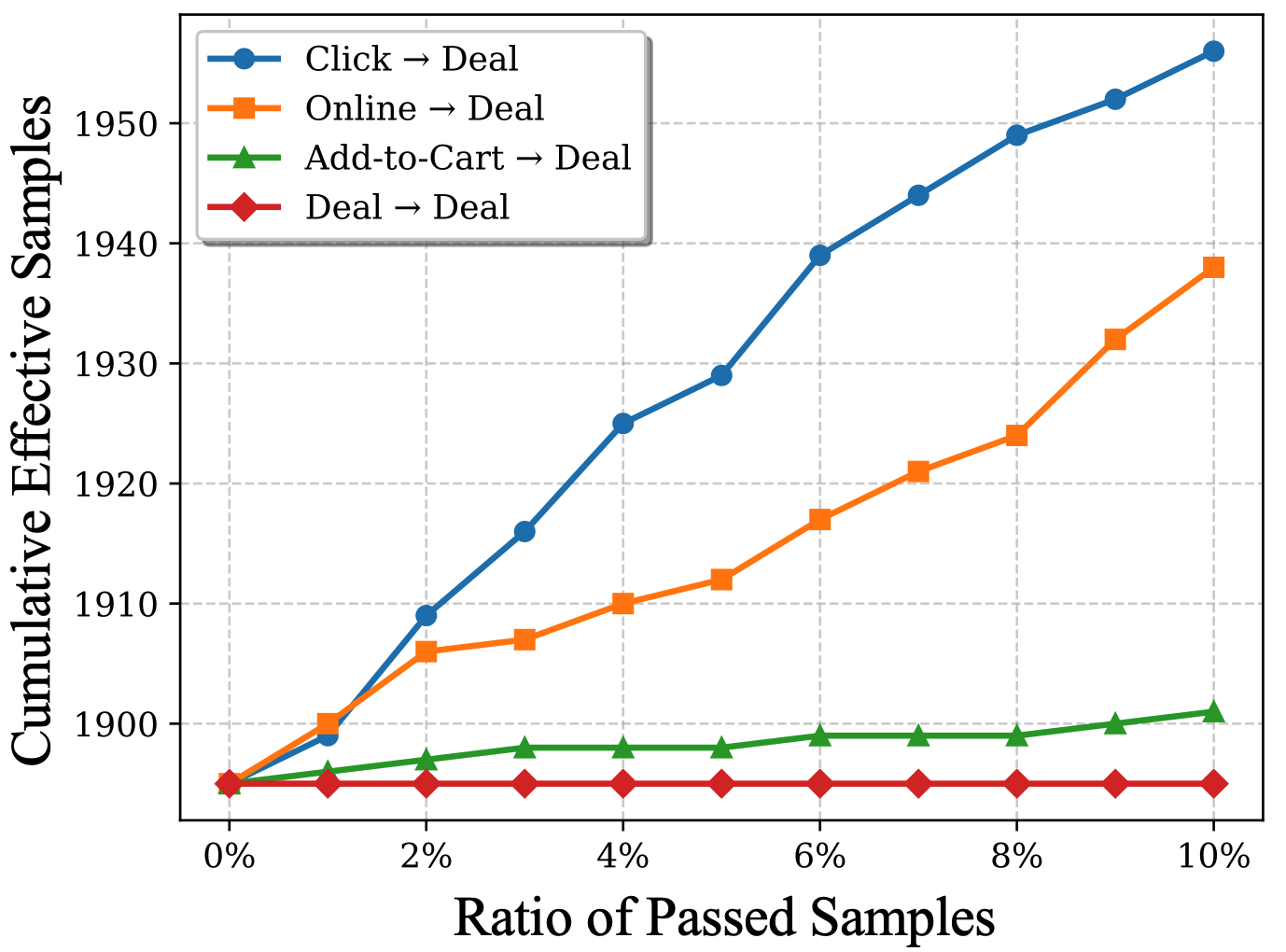}
		\caption{SWAG (Industrial)}
		\label{fig:pass_swag}
	\end{subfigure}
	\hfill
	\begin{subfigure}[b]{0.49\columnwidth}
		\centering
		\includegraphics[width=\textwidth]{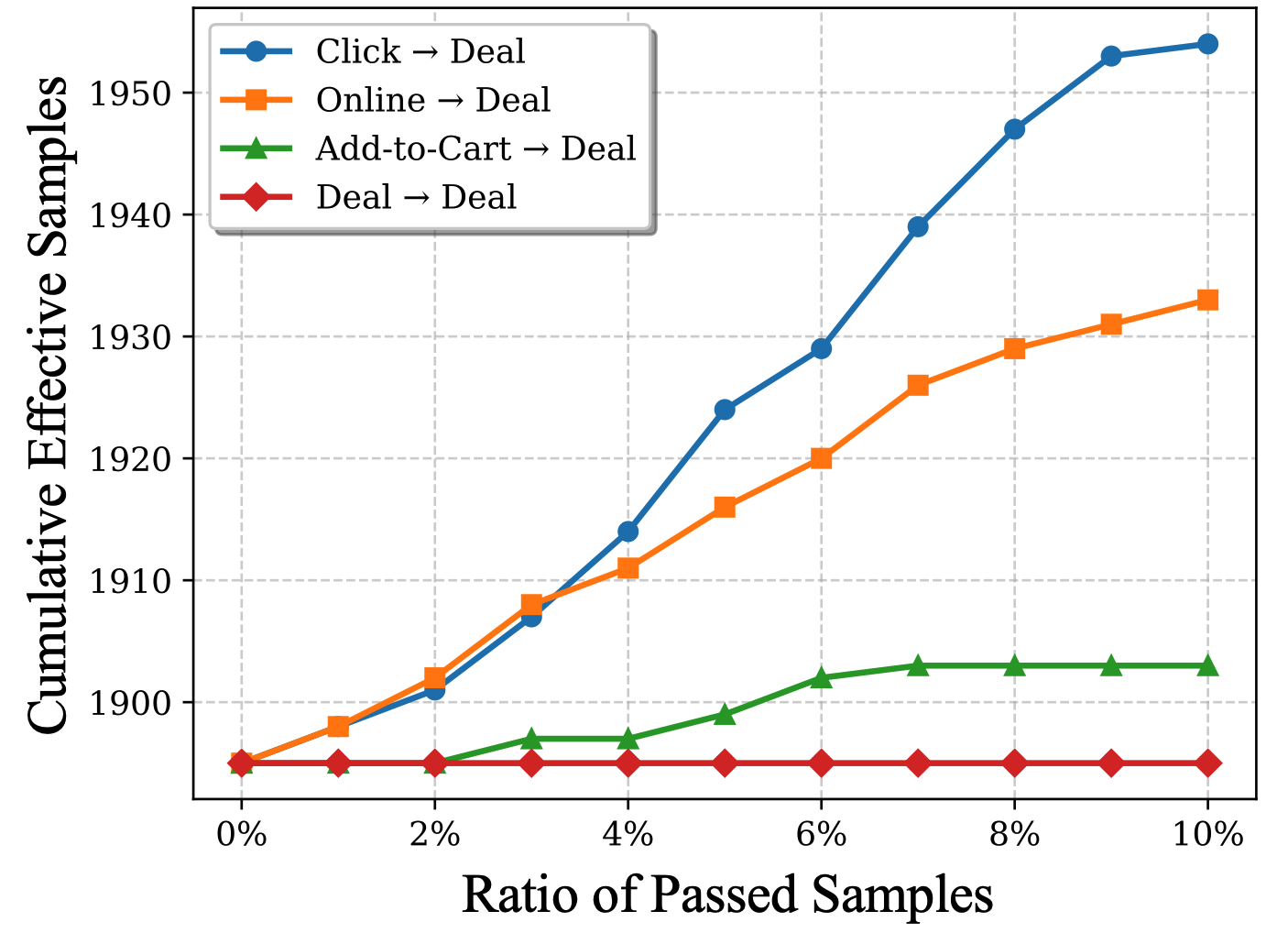}
		\caption{MC Dropout (Industrial)}
		\label{fig:pass_mc}
	\end{subfigure}

	\vspace{2pt}
	\begin{subfigure}[b]{0.49\columnwidth}
		\centering
		\includegraphics[width=\textwidth]{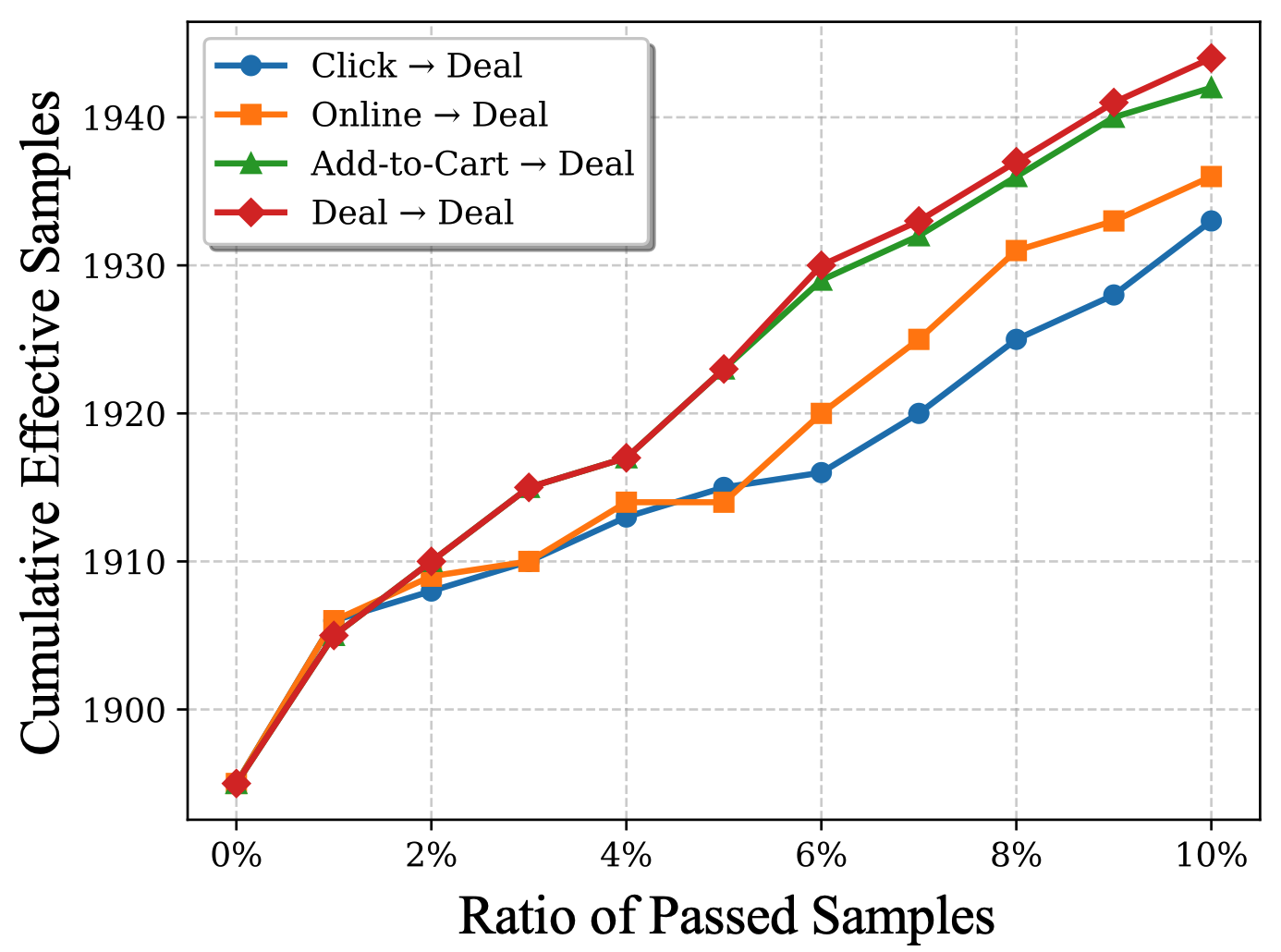}
		\caption{DDU (Industrial)}
		\label{fig:pass_ddu}
	\end{subfigure}
	\hfill
	\begin{subfigure}[b]{0.49\columnwidth}
		\centering
		\includegraphics[width=\textwidth]{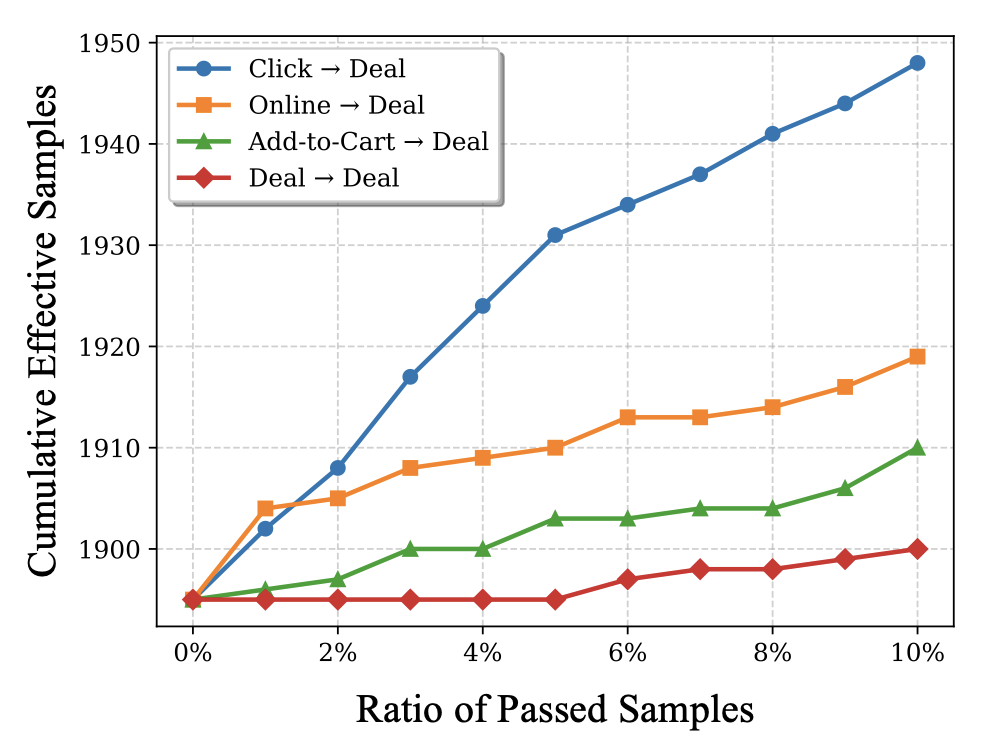}
		\caption{EDL (Industrial)}
		\label{fig:pass_edl}
	\end{subfigure}
	\caption{Evaluation of sample passing capability on the JD dataset. Corresponding results on Criteo are provided in Appendix~\ref{app:criteo}.}
	\label{fig:passing_all}
\end{figure}

Next, we analyze the effective samples supplied to downstream tasks under different uncertainty-based passing strategies. This is the most critical experiment for determining whether our uncertainty-share strategy yields positive returns. The experiments are conducted on the JD dataset and the Criteo dataset. We compare four uncertainty estimation methods: SWAG, MC Dropout, DDU, and EDL. With the objective of maximizing the true deal samples, we evaluate each passing strategy by computing the sum of (i) the true deal samples among the directly passed samples and (ii) the true deal samples correctly classified among the remaining samples. Leveraging the uncertainty-sharing logic of UMDA, we compare the numbers of deal samples obtained by passing based on uncertainties derived from different metrics. The results are presented in Figure~\ref{fig:passing_all}.

Among the four uncertainty estimation methods, SWAG, MC Dropout, and EDL exhibit nearly identical behavior: the uncertainty of the sparse metric deal itself can hardly bring additional true deal samples to the downstream task, whereas the click metric, through uncertainty sharing, delivers more true deal samples downstream. DDU, in contrast, demonstrates a different property. By modeling uncertainty at the class level, DDU differs from uncertainty modeling approaches derived from model design and inference; it possesses inherent robustness to data imbalance. However, this class-based modeling also depends on the distribution of the minority class and is therefore less stable. In the RTA scenario, the number of true deal samples obtained by passing based on DDU uncertainty is generally lower than that achieved by the other uncertainty methods via uncertainty sharing, highlighting the superiority of the uncertainty-sharing mechanism. Consistent trends on the Criteo dataset are reported in Appendix~\ref{app:criteo}.

We also examine the AUC-ROC and AUC-PR of the remaining samples after passing, which indirectly reflect the quality of the passed traffic: once the samples the model is uncertain about are routed downstream, the residual set should become easier to classify. On the balanced C2S click metric this is exactly what we observe (Table~\ref{tab:passing_click}): as the passing ratio grows, the residual AUC-ROC rises from 0.8097 to 0.8290 and AUC-PR from 0.8642 to 0.8880, confirming that the passed traffic is concentrated on the samples the model is genuinely uncertain about.

\begin{table}[htbp]
	\centering
	\caption{Residual classification on the C2S click metric under different passing ratios.}
	\label{tab:passing_click}
	\renewcommand{\arraystretch}{1.15}
	\begin{tabular*}{\columnwidth}{@{\extracolsep{\fill}}lcc@{}}
		\toprule
		Passing ratio & AUC-ROC & AUC-PR \\
		\midrule
		0.20 & \textbf{0.8290} & \textbf{0.8880} \\
		0.15 & 0.8245 & 0.8820 \\
		0.10 & 0.8196 & 0.8760 \\
		0.05 & 0.8146 & 0.8701 \\
		0    & 0.8097 & 0.8642 \\
		\bottomrule
	\end{tabular*}
\end{table}

For the sparse deal metric the picture is more nuanced: passing by deal's own uncertainty does not help, whereas sharing the C2S click uncertainty recovers the loss and yields the best residual performance, echoing the sample-level gains in Figure~\ref{fig:passing_all}. Because AUC-PR is sensitive to the positive-sample ratio, its absolute level on the sparse metric is low and what matters is the relative ordering across strategies; the full nine-column comparison is deferred to Appendix~\ref{app:residual}.

\subsection{Model distillation}
Having introduced the distillation framework in Section~\ref{sec:distill}, we now evaluate whether the distilled single-pass model preserves the uncertainty behavior of the original UMDA. We measure the consistency between the uncertainty measures of the distilled and original models---the degree to which they agree on the \emph{ranking} of uncertainties, which underpins interception regardless of whether high or low uncertainty is favored. We evaluate this ranking consistency for the four metrics on the JD dataset, separately for the high- and low-uncertainty portions (Appendix~\ref{app:distill_fidelity}, Figure~\ref{fig:overlap_8}). For the C2S Click metric, the overlap between the original and distilled models exceeds 80\% for the 10\% of samples with the lowest uncertainty and approaches 60\% for the 10\% with the highest uncertainty---far above the 10\% overlap expected by random choice, indicating that the distilled model has effectively learned the uncertainty modeling. In sparser conversion scenarios, fidelity is somewhat lower, but the distilled model still preserves the original ranking, and distilling uncertainty for these sparse metrics is not our primary target. We further examine how the number of effective deal samples passed to downstream tasks changes after direct and indirect passing (Appendix~\ref{app:distill_fidelity}, Figure~\ref{fig:distill_overall_performance}).

As the passing ratio increases, the uncertainty from C2S click and online remains effective for passing the deal metric, and the overall trend matches that before distillation, so the distilled model preserves the passing capability of the original. Under a batch size of 512 and 11 inference iterations for the original uncertainty model, the distilled model's batch inference time is 9.86\% of the original. This single-pass cost is what allows the full traffic to be scored in the near-line pipeline (Section~\ref{sec:deployment}), and it is the distilled model, not the original UMDA, that we run in production; all online results in Section~\ref{sec:deployment} use this model.

\section{Online Deployment and A/B Test}
\label{sec:deployment}
We deployed the distilled UMDA model in JD.com's production advertising system and evaluated it with an online A/B test on 5\% of live traffic. After the test confirmed its benefits, we rolled the model out to full production traffic.

\begin{figure*}[t]
	\centering
	\scalebox{0.85}{%
	\begin{tikzpicture}[
		font=\small,
		box/.style={draw, semithick, rounded corners, minimum height=9mm, minimum width=22mm, align=center, inner sep=2pt, fill=blue!6},
		nl/.style={draw, semithick, rounded corners, minimum height=9mm, minimum width=22mm, align=center, inner sep=2pt, fill=teal!8},
		on/.style={draw, semithick, rounded corners, minimum height=9mm, minimum width=20mm, align=center, inner sep=2pt, fill=orange!10},
		term/.style={draw, semithick, rounded corners, minimum height=8mm, minimum width=20mm, align=center, inner sep=2pt},
		dec/.style={draw, semithick, diamond, aspect=2.2, align=center, inner sep=0pt, fill=orange!16, minimum width=17mm},
		flow/.style={-{Latex[length=2mm]}, semithick},
		distill/.style={-{Latex[length=2mm]}, semithick, dashed},
		lane/.style={rounded corners, inner sep=3.5mm},
	]
		\node[box] (logs) {Impression \&\\conversion logs};
		\node[box, right=8mm of logs] (ple) {PLE\\multi-objective};
		\node[box, right=8mm of ple] (swag) {SWAG teacher\\($\sim$11 passes)};
		\node[box, right=8mm of swag] (distill) {Uncertainty\\distillation};
		\draw[flow] (logs) -- (ple);
		\draw[flow] (ple) -- (swag);
		\draw[flow] (swag) -- (distill);

		\node[nl, below=13mm of logs] (feat) {Batch\\features};
		\node[nl, below=13mm of ple] (student) {Distilled student\\(single pass)};
		\node[nl, below=13mm of swag] (share) {Shared\\scoring $v_i$};
		\node[dec, below=12mm of distill] (rank) {Top-$(1{-}r)$};
		\node[nl, right=9mm of rank] (bl) {Blacklist\\(hourly)};
		\draw[flow] (feat) -- (student);
		\draw[flow] (student) -- (share);
		\draw[flow] (share) -- (rank);
		\draw[flow] (rank) -- (bl);
		\draw[distill] (distill) -- node[right, font=\scriptsize] {weights} (student);

		\node[on, below=13mm of feat] (req) {RTA\\request};
		\node[dec, right=9mm of req] (hit) {In\\blacklist?};
		\node[term, fill=green!8, right=13mm of hit] (pass) {Pass to\\downstream ads};
		\node[term, fill=red!7, below=6mm of pass] (block) {Intercept};
		\draw[flow] (req) -- (hit);
		\draw[flow] (hit) -- node[above, font=\scriptsize] {no} (pass);
		\draw[flow] (hit) |- node[pos=0.25, left, font=\scriptsize] {yes} (block);
		\draw[flow, densely dotted] (bl) |- node[pos=0.75, above, font=\scriptsize] {lookup} (hit.north);

		\begin{scope}[on background layer]
			\node[lane, fill=blue!3, fit=(logs)(distill)] (laneA) {};
			\node[lane, fill=teal!4, fit=(feat)(bl)] (laneB) {};
			\node[lane, fill=orange!4, fit=(req)(pass)(block)] (laneC) {};
		\end{scope}
		\node[anchor=west, font=\footnotesize\itshape, text=blue!55] at (laneA.north west) {Offline training};
		\node[anchor=west, font=\footnotesize\itshape, text=teal!55!black] at (laneB.north west) {Near-line scoring};
		\node[anchor=west, font=\footnotesize\itshape, text=orange!60!black] at (laneC.north west) {Online lookup (serving path)};
	\end{tikzpicture}%
	}
	\caption{End-to-end architecture of the deployed system. Training and uncertainty distillation run offline, the distilled student scores traffic in a near-line batch pipeline that maintains an hourly blacklist, and the online serving path performs only a lightweight blacklist lookup without model inference.}
	\label{fig:system}
\end{figure*}

\subsection{System architecture}
Figure~\ref{fig:system} shows UMDA in the production pipeline. Training is offline: the PLE backbone learns four funnel objectives, the SWAG teacher produces multi-pass uncertainty, and distillation (Section~\ref{sec:distill}) compresses them into a single-pass student. The student scores traffic in a near-line batch pipeline, emitting probability and uncertainty per objective per request, which the funnel-shared scoring rule (Section~\ref{sec:approx}) combines into $v_i$. The lowest $(1{-}r)$ fraction is blacklisted (refreshed hourly), while highest-uncertainty traffic is deliberately exempted to pass downstream. At serving time, only a lightweight blacklist lookup occurs---matched requests are intercepted, others pass to ads. All model inference stays off the latency-critical path.

\subsection{Deployment setting and scale}
The system operates at industrial scale: off-site advertising in JD.com's system generates on the order of hundreds of billions of requests per day, and the RTA channel alone serves up to hundreds of millions of requests per minute. The original UMDA requires roughly eleven stochastic forward passes per request (Section~\ref{sec:distill}), so scoring the full traffic with the teacher would exceed the near-line compute budget. Distillation reduces this to a single forward pass, at 9.86\% of the teacher's batch inference cost, which brings full-traffic scoring within budget. We therefore run only the distilled model in production, and all results below are obtained with it.

\subsection{A/B test protocol}
We ran a randomized A/B test on 5\% of live traffic over seven consecutive days. The control group used the existing production interception strategy and the treatment group used our method, both under a matched interception ratio. All reported numbers are traffic-normalized relative differences between treatment and control. We track two groups of metrics. \emph{Primary metrics} measure interception quality: the click fraud rate, CVR, and customer acquisition cost (CAC). \emph{Guardrail metrics} monitor overall traffic and business volume to ensure the interception does not harm healthy traffic: daily converted users (DAC), click count, and daily active users (DAU).

\subsection{Online results and analysis}
Table~\ref{tab:ab_overall} summarizes the seven-day averaged results; the day-by-day breakdown is reported in Appendix~\ref{app:daily} (Table~\ref{tab:ab_daily}). The primary metrics confirm the intended effect. The click fraud rate drops by 3.59\% and CVR improves by 4.01\%, indicating that the intercepted traffic is largely low-quality while the passed traffic is of higher quality, and CAC decreases by 0.26\%. Because our method reduces the volume of passed low-quality requests by design, the guardrail metrics also decline: click count by 3.78\% and DAU by 4.11\%. The number of converted users (DAC) stays essentially unchanged ($-0.72\%$) while conversion quality improves, so the drop in clicks and active users falls on low-quality and fraudulent traffic rather than genuine converters. We view the small DAC decrease as an acceptable tradeoff for the lower fraud rate, higher conversion quality, and reduced acquisition cost. The day-by-day breakdown in Appendix~\ref{app:daily} shows these effects are stable across all seven days.

\begin{table}[htbp]
	\centering
	\caption{Seven-day averaged online A/B results.}
	\label{tab:ab_overall}
	\setlength{\tabcolsep}{11pt}
	\renewcommand{\arraystretch}{1.15}
	\begin{tabular}{llc}
		\toprule
		Type & Metric & Relative change \\
		\midrule
		\multirow{3}{*}{Primary}
		& Click fraud rate & $-3.59\%$ \\
		& CVR              & $+4.01\%$ \\
		& CAC              & $-0.26\%$ \\
		\midrule
		\multirow{3}{*}{Guardrail}
		& DAC (converted users) & $-0.72\%$ \\
		& Click count           & $-3.78\%$ \\
		& DAU                   & $-4.11\%$ \\
		\bottomrule
	\end{tabular}
\end{table}

\subsection{Lessons learned}
Deploying uncertainty-aware interception at industrial scale surfaced three practical lessons.

\noindent\textbf{Treat serving cost as a first-class constraint.} The original UMDA requires about eleven forward passes per request, exceeding the near-line budget for hundreds of billions of daily requests. Distilling uncertainty into a single pass at 9.86\% of the teacher's cost made full-traffic scoring practical, so we designed for the serving budget from the outset rather than retrofitting afterwards.

\noindent\textbf{Share uncertainty from where it is trustworthy.} Intercepting each objective by its own uncertainty fails for sparse objectives like deal, whose uncertainty collapses onto predicted probability (Section~\ref{sec:approx}). Routing deal traffic by the balanced click metric's uncertainty—shared along the funnel—makes the signal usable. For the same reason, interception needs only a \emph{relative} ordering of traffic, so rescaling the distillation target (Section~\ref{sec:distill}) lets the student learn this ordering without the classification loss dominating.

\noindent\textbf{Read guardrails together, not in isolation.} A 4\% drop in DAU looks alarming on its own, but read alongside a lower fraud rate, higher CVR, and stable converted users, it indicates that the removed traffic was mostly low-quality. Monitoring primary and guardrail metrics jointly was essential during rollout.

\section{Conclusion}
We analyzed why weight-based uncertainty degrades under label imbalance, and proposed UMDA, which shares uncertainty across funnel-correlated objectives to overcome this limitation, together with a distilled variant that preserves this capability at roughly one-tenth the inference cost. This speedup is what lets the full traffic be scored in a near-line pipeline, where the multi-pass teacher would not fit the compute budget. Deployed at JD.com to maintain the interception blacklist, UMDA reduced the click fraud rate by 3.59\% and improved CVR by 4.01\% in a seven-day online A/B test while keeping converted users essentially unchanged, and has since been rolled out to full production traffic. Future work could exploit the interrelationships among the multiple uncertainties produced by multi-objective learning to further improve estimation quality in sparse-label settings, for example, by jointly calibrating uncertainty estimates across related tasks to better handle data sparsity.

\bibliographystyle{ACM-Reference-Format}
\bibliography{example_paper}

\appendix

\section{Verification of Uncertainty Properties}
\label{app:uncertainty_props}
\begin{figure}[htbp]
	\centering
	\begin{subfigure}[b]{0.46\columnwidth}
		\centering
		\includegraphics[width=\textwidth]{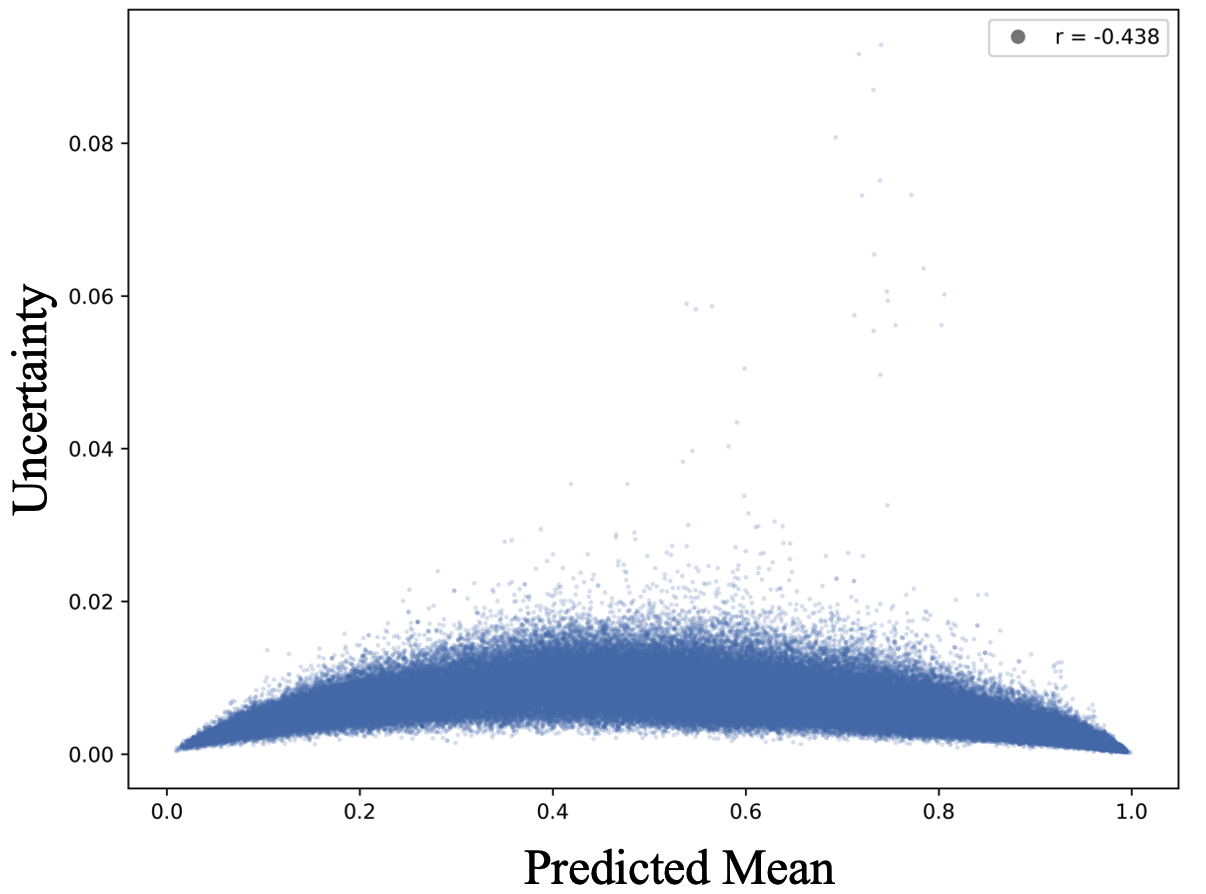}
		\caption{C2S click}
		\label{fig:uncertainty_c2s}
	\end{subfigure}
	\hfill
	\begin{subfigure}[b]{0.46\columnwidth}
		\centering
		\includegraphics[width=\textwidth]{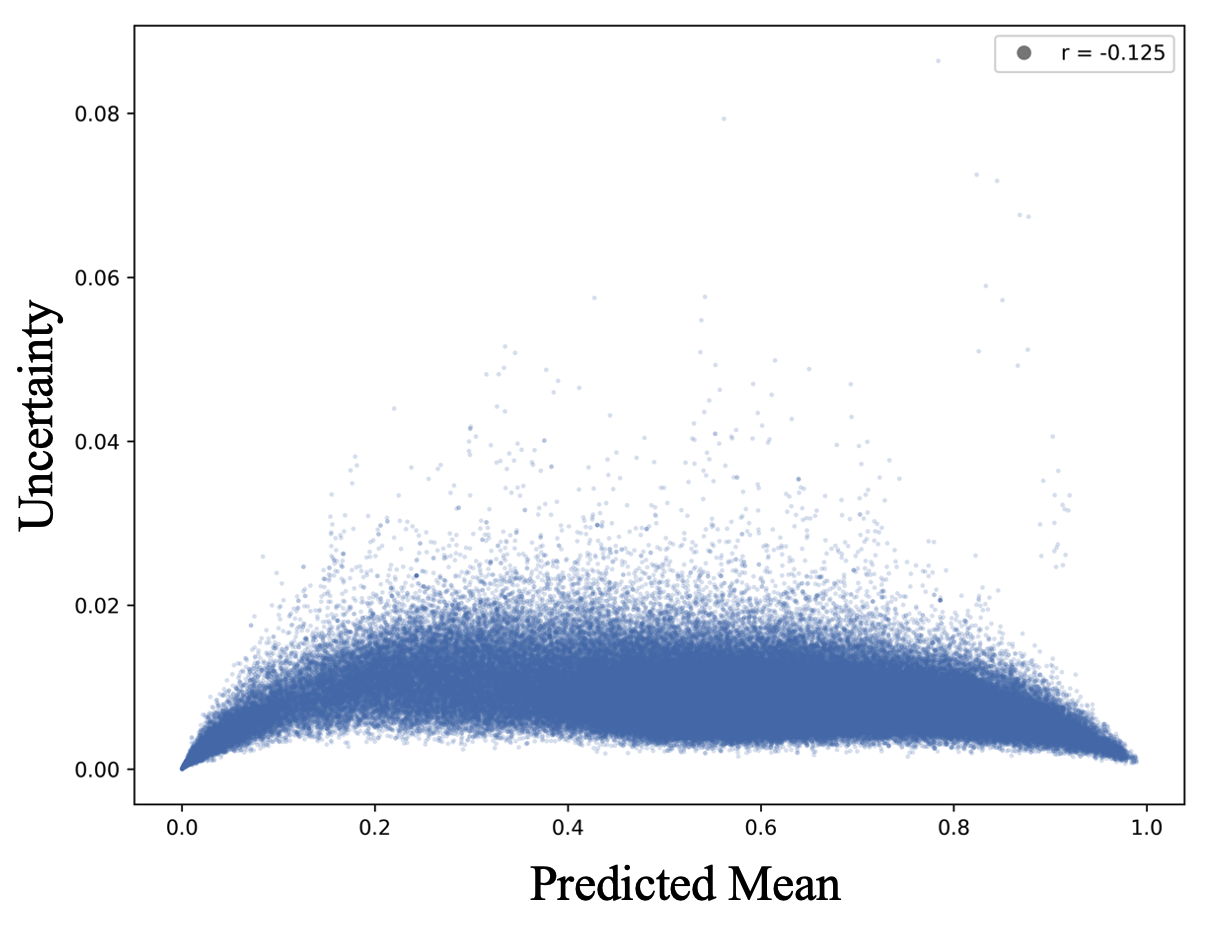}
		\caption{Online}
		\label{fig:uncertainty_online}
	\end{subfigure}

	\vspace{2pt}
	\begin{subfigure}[b]{0.46\columnwidth}
		\centering
		\includegraphics[width=\textwidth]{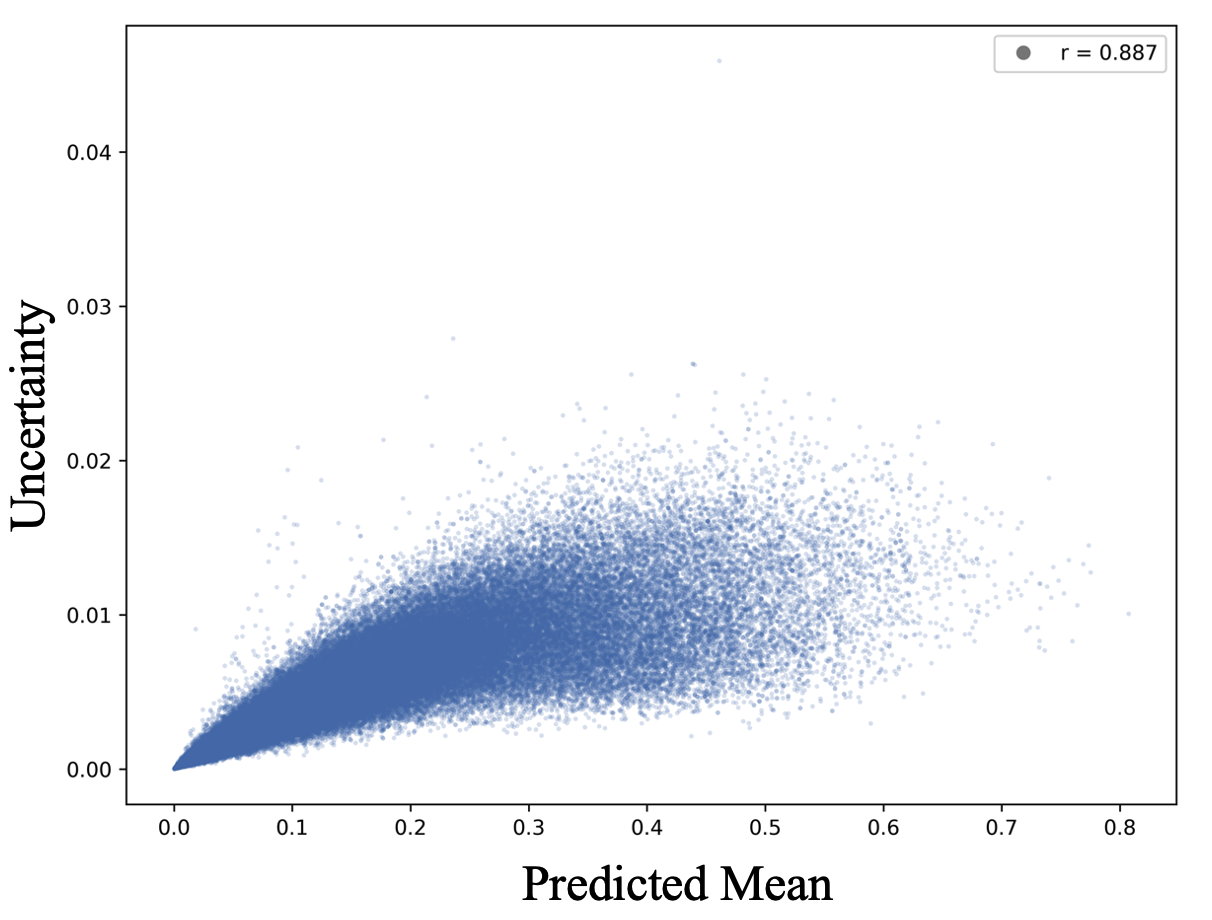}
		\caption{Add to cart}
		\label{fig:uncertainty_cart}
	\end{subfigure}
	\hfill
	\begin{subfigure}[b]{0.46\columnwidth}
		\centering
		\includegraphics[width=\textwidth]{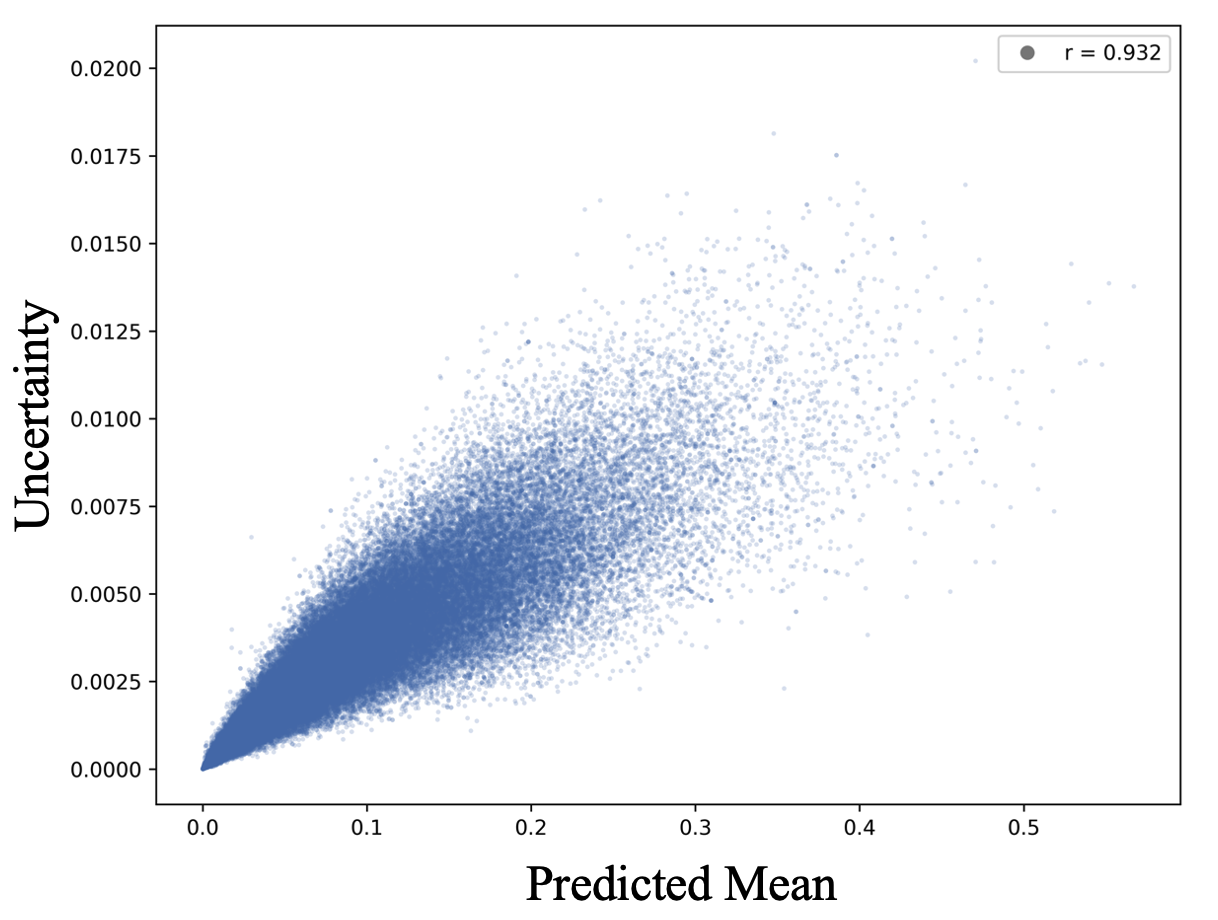}
		\caption{Deal}
		\label{fig:uncertainty_deal}
	\end{subfigure}
	\caption{Relationship between predicted probability and uncertainty for four behavioral metrics.}
	\label{fig:uncertainty_all}
\end{figure}

Figure~\ref{fig:uncertainty_all} shows the relationship between predicted probability and the estimated uncertainty for each of the four behavioral metrics, supporting the analysis in Section~\ref{sec:exp}. For the balanced C2S click and online metrics, uncertainty peaks near a predicted probability of 0.5 and decays symmetrically toward both extremes, matching the theoretical property that epistemic uncertainty maximizes at the decision boundary. For the sparse add-to-cart and deal metrics, uncertainty instead increases monotonically with the predicted probability (Pearson correlation 0.887 and 0.932, respectively), showing that under extreme sparsity the uncertainty degenerates into a copy of the confidence score.

\section{Full Derivation of the Uncertainty Expression}
\label{app:derivation}
This appendix provides the complete derivation of the closed-form uncertainty expression~\eqref{uncertainty_expression1} and the neighbor-influence term~\eqref{neighbor1} stated in Section~\ref{sec:approx}.

\subsection{Prediction dynamics under SGD}
We analyze how the prediction changes as the model weights are updated via stochastic gradient descent (SGD). The SGD update rule is
\begin{equation}\label{app:SGD}
	w^{(t+1)} = w^{(t)} - \eta \nabla_w \ell(w^{(t)}; x^*, y),
\end{equation}
where $\eta$ is the learning rate and $\ell$ denotes the loss function. For a binary classification task with sigmoid activation $\sigma(\cdot)$, the gradient of the loss with respect to the weights is
\[
\nabla_w \ell(w^{(t)}; x^*, y) = \bigl(\sigma(f_{w^{(t)}}(x^*)) - y\bigr) \,\nabla_w f_{w^{(t)}}(x^*).
\]
Define $c_t = \|\nabla_w f_{w^{(t)}}(x^*)\|^2 > 0$ as the squared gradient norm at step $t$; when the model architecture is fixed, $c_t$ is a constant determined by $w^{(t)}$. A first-order Taylor expansion with remainder gives
\begin{align*}
	f_{w^{(t+1)}}(x^*)
	&= f_{w^{(t)}}(x^*) + \nabla_w f_{w^{(t)}}(x^*)^\top\!\bigl(w^{(t+1)} - w^{(t)}\bigr) \\
	&\qquad + \mathcal{O}\bigl(\|w^{(t+1)}-w^{(t)}\|^2\bigr).
\end{align*}
Substituting the SGD update \eqref{app:SGD} and neglecting the higher-order terms, we obtain the approximate prediction update
\begin{equation}
	f_{w^{(t+1)}}(x^*)
	\approx f_{w^{(t)}}(x^*) - \eta \,c_t \bigl(\sigma(f_{w^{(t)}}(x^*)) - y \bigr).
	\label{eq:pred_update}
\end{equation}
This approximation is generally reasonable when the learning rate is small and the magnitude of weight updates is small.

\subsection{Error dynamics as an AR(1) process}
Let $f^*$ satisfy $\sigma(f^*) = q$; $f^*$ maps features to the corresponding conditional probabilities, representing the idealized optimal network. Let $e_t = f_{w^{(t)}}(x^*) - f^*$ be the prediction error. Linearizing $\sigma$ around $f^*$ gives
\[
\sigma(f_{w^{(t)}}(x^*)) \approx \sigma(f^*) + \sigma'(f^*)\,e_t = q + q(1-q)e_t,
\]
where we used $\sigma'(f^*) = q(1-q)$. Inserting this into \eqref{eq:pred_update} yields
\[
f_{w^{(t+1)}}(x^*)
\approx f^* + \bigl[1 - \eta c_t q (1-q)\bigr] e_t + \eta c_t (y - q),
\]
so the error evolves according to
\begin{equation}
	\label{eq:error_dynamics}
	e_{t+1} \approx \bigl[1 - \eta c_t q (1-q)\bigr] e_t + \eta c_t (y - q).
\end{equation}
Under the assumption that the model has converged well and the weight updates are slow, $c_t$ varies slowly and can be approximated by a constant $c$. The error dynamics then reduce to an AR(1) process $e_{t+1} = \alpha e_t + \epsilon_t$, with $\alpha = 1 - \eta c q (1-q)$ and innovation $\epsilon_t = \eta c (y-q)$. For $|\alpha| < 1$, the stationary variance is $\mathrm{Var}(e_t) = \mathrm{Var}(\epsilon_t)/(1 - \alpha^2)$. Computing the terms explicitly,
\[
\mathrm{Var}(\epsilon_t) = (\eta c)^2 \,\mathrm{Var}(y) = (\eta c)^2 q(1-q),
\]
\begin{align}
	1 - \alpha^2 &= 1 - \bigl[1 - \eta c q(1-q)\bigr]^2 \notag\\
	&= \eta c q(1-q)\bigl[2 - \eta c q(1-q)\bigr].
	\label{eq:denominator}
\end{align}
Substituting these into the variance formula yields the uncertainty measure of \eqref{uncertainty_expression1}:
\[
\mathrm{Var}(e_t) = \frac{(\eta c)^2 q(1-q)}{\eta c q(1-q)\,[2 - \eta c q(1-q)]}
= \frac{\eta c}{2 - \eta c q (1-q)}.
\]

\subsection{Influence of neighboring samples}
To characterize how $c$ depends on nearby samples, we focus on the output layer, $q(x;w) = \sigma(w^\top \phi(x))$. Assume that for unseen similar samples, the features passed to the output layer remain close to each other. For a sample $x_1$, the predicted probability is $q_1 := q(x_1; w)$ and the feature representation is $\phi_1 := \phi(x_1)$. The gradient and Hessian of the output at $x_1$ are
\[
g_1 := \nabla_w q(x_1;w) = s_1 \phi_1, \qquad s_1 := q_1(1-q_1),
\]
\begin{equation}
	H_1 := \nabla_w^2 q(x_1;w) = s_1(1-2q_1)\phi_1\phi_1^\top.
\end{equation}
Now consider a neighboring sample $x_2$ used for a parameter update, with true conditional probability $p_2 := p(y=1 \mid x_2)$. Because $x_2$ is close to $x_1$ in feature space, the model's initial prediction for $x_2$ is approximately $q_1$. The cross-entropy loss for $x_2$, averaged over the label distribution, is
\begin{align*}
	\mathbb{E}\bigl[\ell(y, q(x_2;w))\bigr]
	= -\bigl[ & p_2 \log q(x_2;w) + \\
	& (1-p_2)\log(1-q(x_2;w)) \bigr].
\end{align*}
Taking the gradient with respect to $w$ and evaluating at $q(x_2;w) \approx q_1$ gives $\nabla_w \mathbb{E}[\ell] \approx (q_1 - p_2)\,\phi_2$. A gradient descent step with learning rate $\eta$ thus changes the weights by
\begin{equation}
	\Delta w \approx -\eta \,(q_1 - p_2)\,\phi_2.
	\label{eq:delta_w}
\end{equation}
Note that this differs from the gradient of the predicted probability, which would involve an additional factor $s_2 = q_2(1-q_2)$; because we use the expected log-likelihood gradient, that factor does not appear. The resulting change in the gradient at $x_1$ is estimated using the Hessian:
\begin{align}
	\Delta g_1 &= H_1 \Delta w \notag\\
	&\approx s_1 (1-2q_1)\,\phi_1\phi_1^\top \bigl[-\eta (q_1 - p_2)\,\phi_2\bigr] \notag\\
	&= -\eta\, s_1 (1-2q_1)(q_1 - p_2)\,(\phi_1^\top \phi_2)\,\phi_1.
	\label{eq:delta_g}
\end{align}
The relative change in the gradient norm—which directly modulates $c$ and hence the uncertainty—is therefore proportional to $(1 - 2q_1)(p_2 - q_1)\, \phi_1^\top \phi_2$, which is \eqref{neighbor1}.

\section{Distillation Fidelity}
\label{app:distill_fidelity}
\begin{figure}[htbp]
	\centering
	\begin{subfigure}[b]{0.46\columnwidth}
		\centering
		\includegraphics[width=\textwidth]{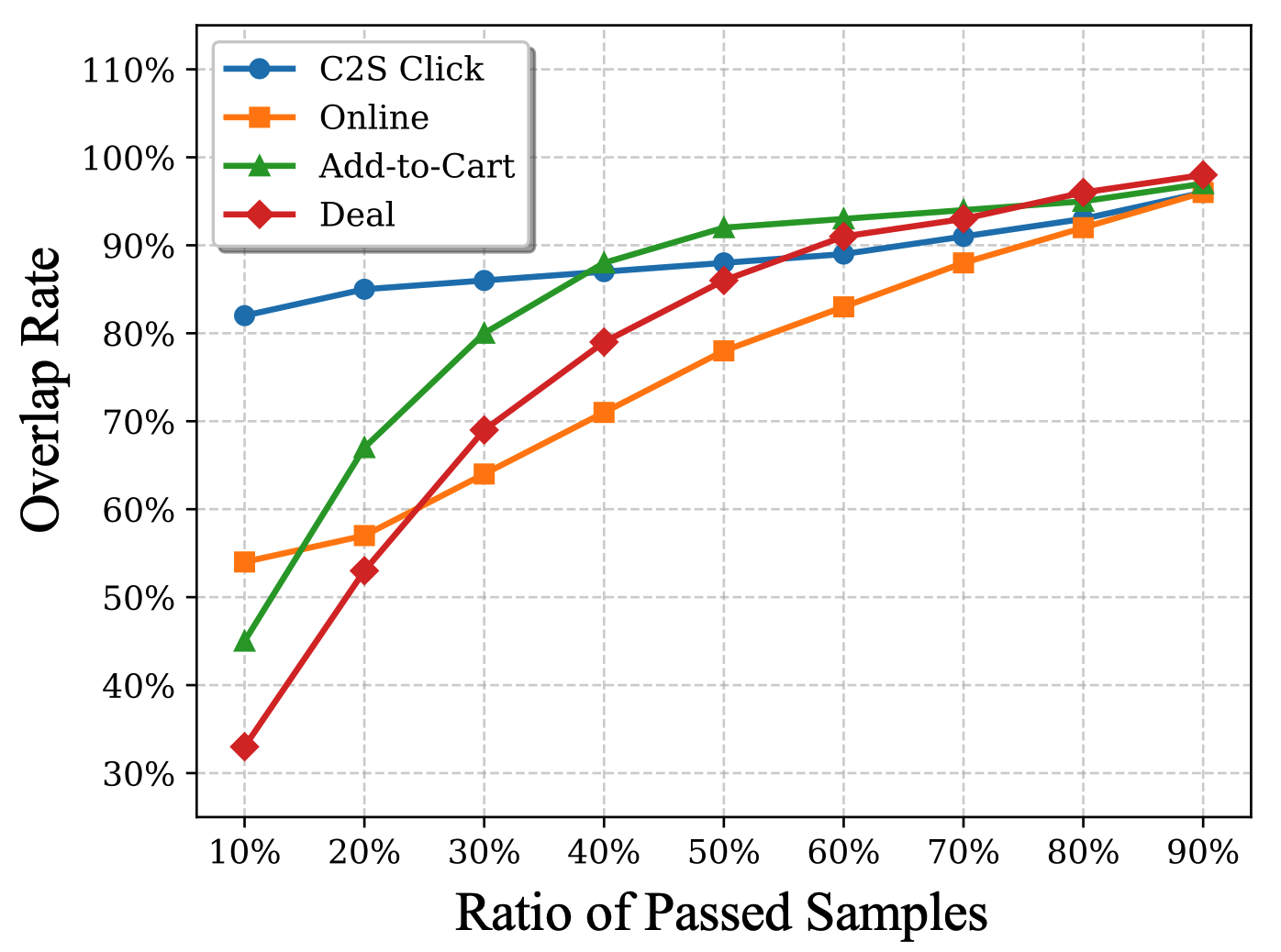}
		\caption{Overlap rate (Low)}
		\label{fig:overlap_low}
	\end{subfigure}
	\hfill
	\begin{subfigure}[b]{0.46\columnwidth}
		\centering
		\includegraphics[width=\textwidth]{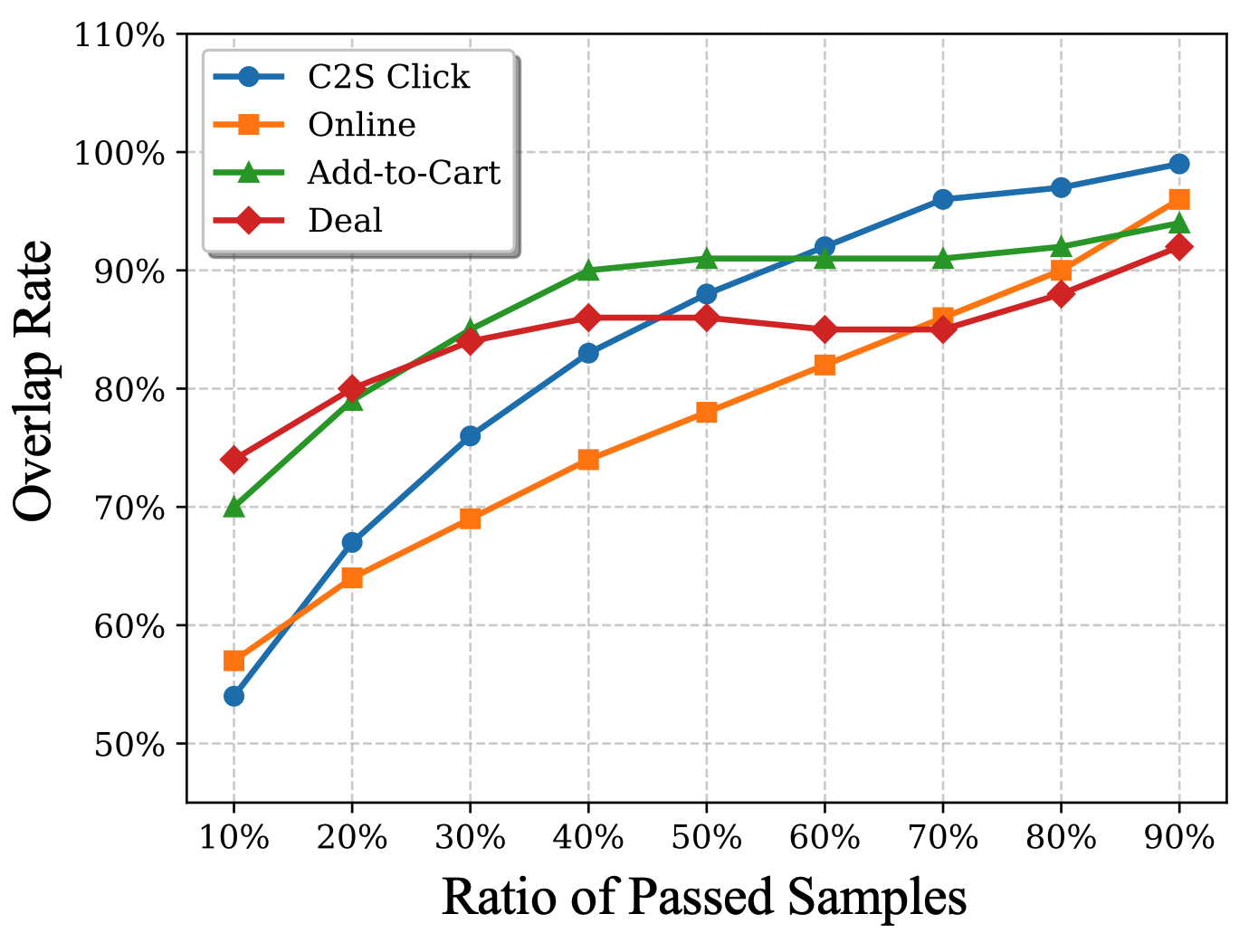}
		\caption{Overlap rate (High)}
		\label{fig:overlap_high}
	\end{subfigure}
	\caption{Overlap ratios between distilled and original model uncertainties for each dimension.}
	\label{fig:overlap_8}
\end{figure}

\begin{figure}[htbp]
	\centering
	\includegraphics[width=0.72\columnwidth]{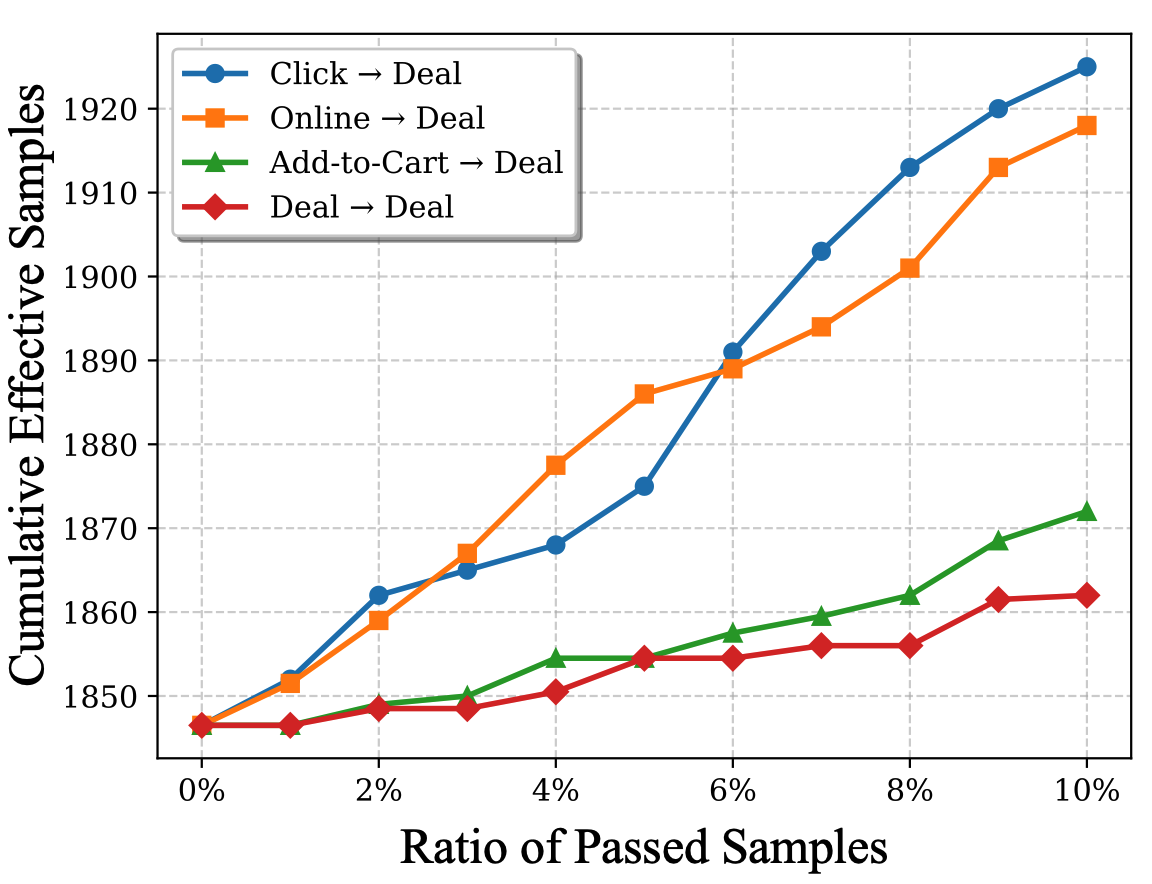}
	\caption{Total number of deal samples passed to downstream under different passing strategies.}
	\label{fig:distill_overall_performance}
\end{figure}

This appendix provides the supporting figures for the distillation-fidelity analysis discussed in Section~\ref{sec:distill}. Figure~\ref{fig:overlap_8} reports the overlap between the uncertainty rankings of the original and distilled models, for the lowest- and highest-uncertainty portions of each metric. Figure~\ref{fig:distill_overall_performance} reports the total number of deal samples passed to downstream tasks under different passing strategies, before and after distillation.

\section{Neighborhood Effect on Uncertainty}
\label{app:neighbor}
\begin{figure}[htbp]
	\centering
	\includegraphics[width=0.72\columnwidth]{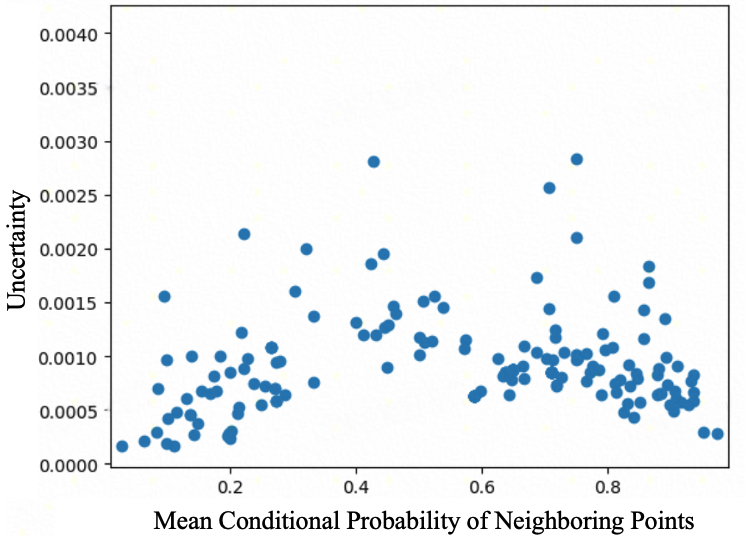}
	\caption{Relationship between epistemic uncertainty and neighborhood conditional probability for test samples with a predicted probability of exactly 0.5 on the C2S click metric. The estimated uncertainty shifts systematically as the mean conditional probability of training samples within a Euclidean distance of 1 deviates from the decision boundary.}
	\label{fig:neighbor_uncertainty}
\end{figure}

To empirically isolate the neighborhood effect predicted by \eqref{neighbor1}, we selected all test samples with conditional probability exactly 0.5 on the C2S click metric, and for each examined all samples within a Euclidean distance of 1 in the input feature space, computing their conditional probabilities. Figure~\ref{fig:neighbor_uncertainty} shows how the estimated uncertainty for these central samples varies with the neighborhood statistic, while the conditional probability itself remains fixed at the decision boundary. As the mean conditional probability of the neighbors deviates from 0.5 in either direction, the estimated uncertainty of the central samples shifts correspondingly, confirming that uncertainty reflects the label consistency of the surrounding training data rather than the pointwise conditional probability alone.

\section{Passing Capability on the Criteo Dataset}
\label{app:criteo}
\begin{figure}[htbp]
	\centering
	\begin{subfigure}[b]{0.46\columnwidth}
		\centering
		\includegraphics[width=\textwidth]{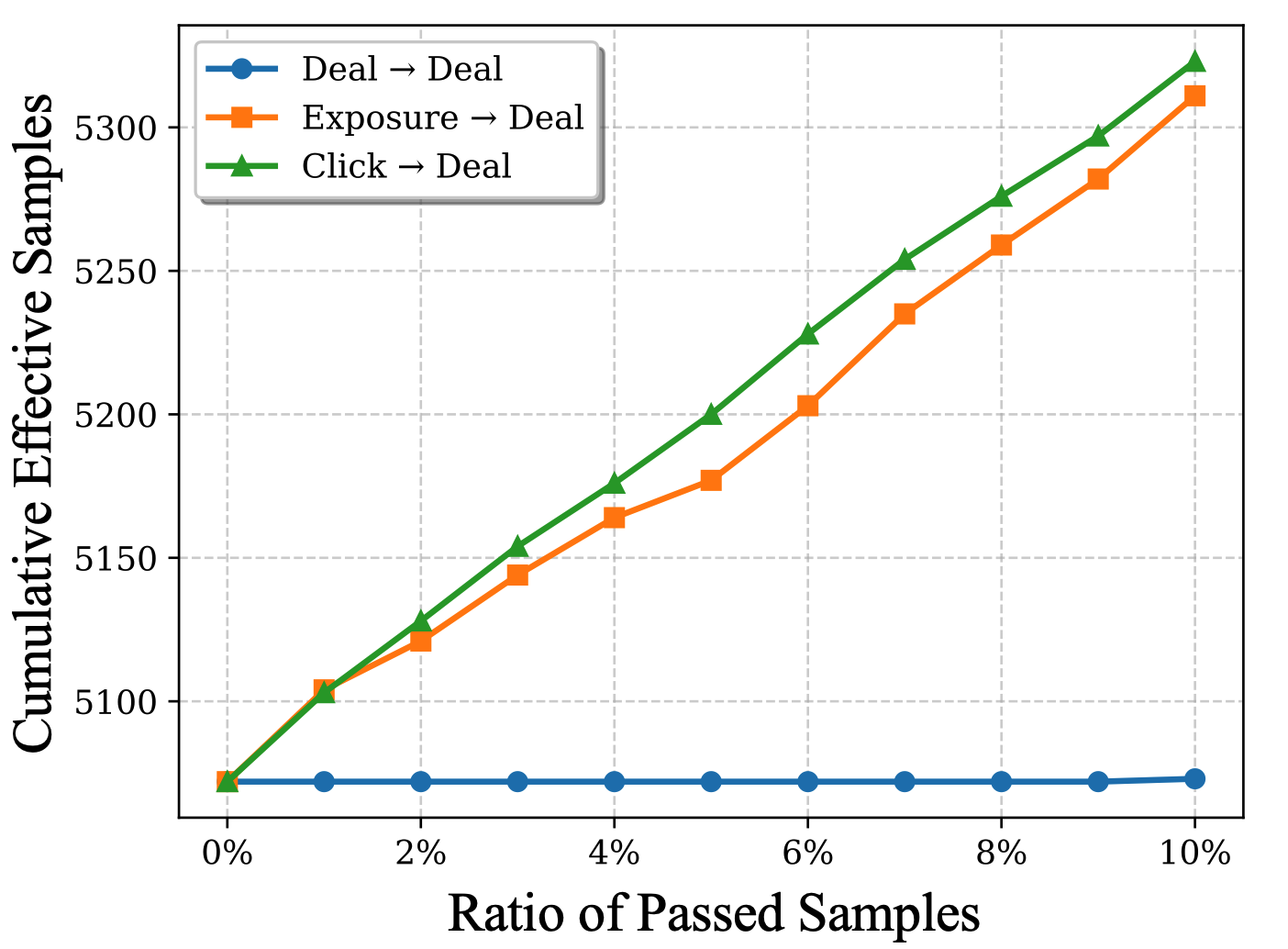}
		\caption{SWAG (Criteo)}
		\label{fig:pass_umda_criteo}
	\end{subfigure}
	\hfill
	\begin{subfigure}[b]{0.46\columnwidth}
		\centering
		\includegraphics[width=\textwidth]{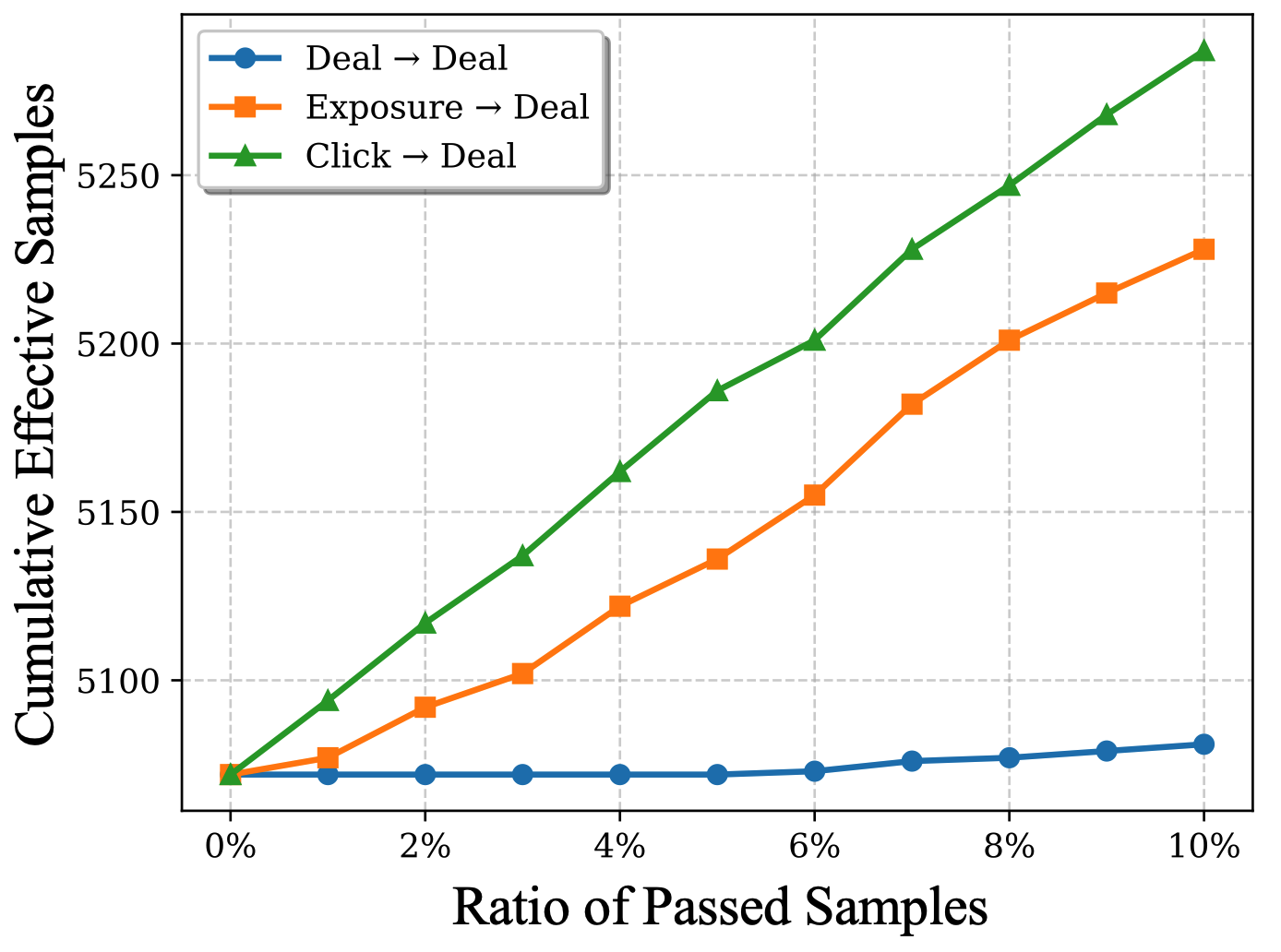}
		\caption{MC Dropout (Criteo)}
		\label{fig:pass_mc_criteo}
	\end{subfigure}

	\vspace{2pt}
	\begin{subfigure}[b]{0.46\columnwidth}
		\centering
		\includegraphics[width=\textwidth]{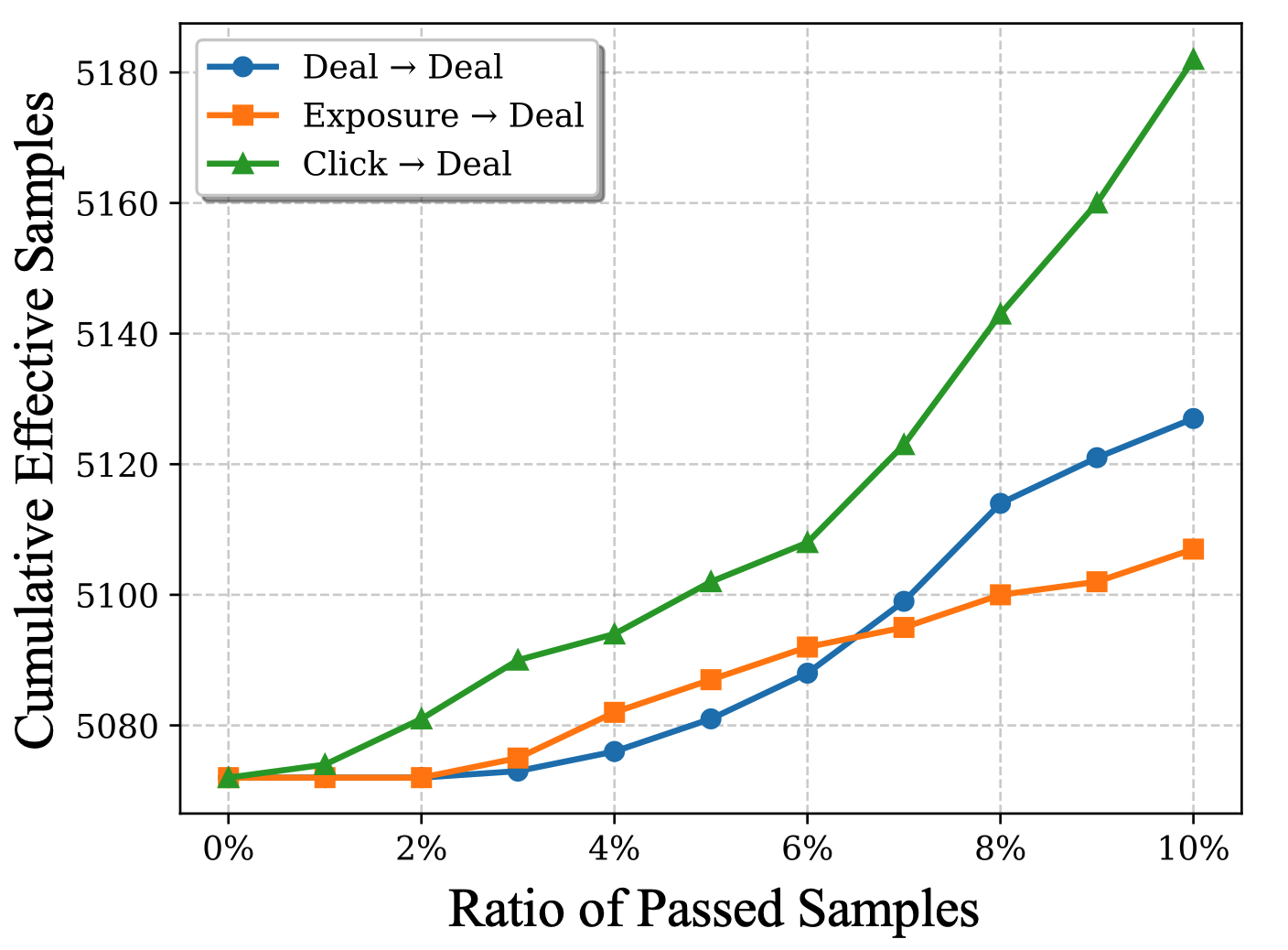}
		\caption{DDU (Criteo)}
		\label{fig:pass_ddu_criteo}
	\end{subfigure}
	\hfill
	\begin{subfigure}[b]{0.46\columnwidth}
		\centering
		\includegraphics[width=\textwidth]{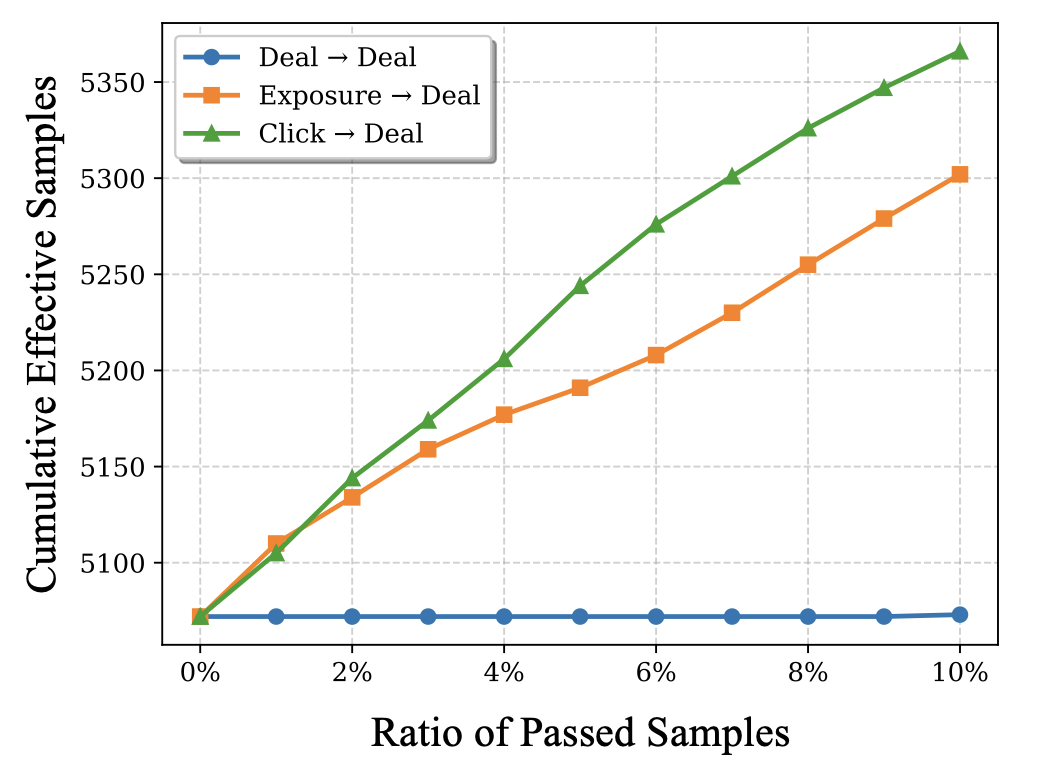}
		\caption{EDL (Criteo)}
		\label{fig:pass_edl_criteo}
	\end{subfigure}
	\caption{Evaluation of sample passing capability on the Criteo dataset.}
	\label{fig:passing_criteo}
\end{figure}

Figure~\ref{fig:passing_criteo} reports the sample passing capability on the Criteo dataset, complementing the industrial results in Figure~\ref{fig:passing_all}.

\section{Day-by-Day Online A/B Results}
\label{app:daily}
Table~\ref{tab:ab_daily} shows the daily results of the seven‑day online A/B test from Section~\ref{sec:deployment}. All metrics are directionally consistent each day: CVR is positive every day, fraud rate drops on six of seven, and guardrail metrics remain stable in magnitude and direction, indicating stable, non‑incidental effects.

\begin{table}[htbp]
	\centering
	\caption{Day-by-day online A/B results (relative change, treatment vs.\ control). The seven-day averages are reported in Table~\ref{tab:ab_overall}.}
	\label{tab:ab_daily}
	\small
	\renewcommand{\arraystretch}{1.15}
	\begin{tabular*}{\columnwidth}{@{\extracolsep{\fill}}lccccc@{}}
		\toprule
		Date & CVR & DAC & DAU & Fraud rate & CAC \\
		\midrule
		Day 1 & $+3.53\%$ & $-0.10\%$ & $-4.14\%$ & $-0.55\%$ & $-0.92\%$ \\
		Day 2 & $+3.71\%$ & $-0.83\%$ & $-4.09\%$ & $-6.67\%$ & $-0.12\%$ \\
		Day 3 & $+3.49\%$ & $-0.33\%$ & $-4.20\%$ & $-7.95\%$ & $-0.37\%$ \\
		Day 4 & $+4.53\%$ & $-0.54\%$ & $-3.99\%$ & $-6.31\%$ & $-0.75\%$ \\
		Day 5 & $+4.77\%$ & $-1.00\%$ & $-4.21\%$ & $+1.92\%$ & $+0.17\%$ \\
		Day 6 & $+4.34\%$ & $-1.24\%$ & $-4.13\%$ & $-5.18\%$ & $-0.64\%$ \\
		Day 7 & $+3.73\%$ & $-1.02\%$ & $-4.05\%$ & $-0.36\%$ & $+0.84\%$ \\
		\bottomrule
	\end{tabular*}
\end{table}

\section{Full Residual-Classification Comparison}
\label{app:residual}
Table~\ref{tab:passing_comparison} reports full residual-classification results from Section~\ref{sec:exp}, extending the C2S click summary (Table~\ref{tab:passing_click}) to all three passing strategies. Passing high-uncertainty C2S traffic improves residual classification on the balanced click metric, but deal uncertainty does not: its weight-based measure relies on a 0.5 boundary that fails under sparsity-induced distribution shift. Routing deal traffic via well-modeled C2S uncertainty recovers this loss and achieves the best residual performance. AUC-PR is sensitive to the positive-sample ratio, so its absolute level on the sparse deal metric drops as the passing ratio increases; the meaningful comparison is relative change. At passing ratio 0.02, C2S click uncertainty improves AUC-ROC by 1.03\% and AUC-PR by 11.87\% over deal uncertainty.
\begin{table*}[htbp]
	\centering
	\caption{Residual classification (AUC-ROC and AUC-PR) under different passing ratios and strategies.}
	\label{tab:passing_comparison}
	\setlength{\tabcolsep}{8pt}
	\renewcommand{\arraystretch}{1.1}
	\begin{tabular}{ccc ccc ccc}
		\toprule
		\multicolumn{3}{c}{C2S click traffic passing} &
		\multicolumn{3}{c}{Deal traffic passing} &
		\multicolumn{3}{c}{C2S click-based passing on deal traffic} \\
		\cmidrule(lr){1-3} \cmidrule(lr){4-6} \cmidrule(lr){7-9}
		Passing ratio & AUC-ROC & AUC-PR &
		Passing ratio & AUC-ROC & AUC-PR &
		Passing ratio & AUC-ROC & AUC-PR \\
		\midrule
		0.20 & \textbf{0.8290} & \textbf{0.8880} &
		0.020 & 0.7890 & 0.1053 &
		0.020 & \textbf{0.7972} & 0.1178 \\
		0.15 & 0.8245 & 0.8820 &
		0.015 & 0.7911 & 0.1088 &
		0.015 & 0.7970 & 0.1186 \\
		0.10 & 0.8196 & 0.8760 &
		0.010 & 0.7926 & 0.1112 &
		0.010 & 0.7966 & 0.1182 \\
		0.05 & 0.8146 & 0.8701 &
		0.005 & 0.7944 & 0.1154 &
		0.005 & 0.7968 & \textbf{0.1192} \\
		0    & 0.8097 & 0.8642 &
		0     & \textbf{0.7962} & \textbf{0.1191} &
		0     & 0.7962 & 0.1191 \\
		\bottomrule
	\end{tabular}
\end{table*}

\end{document}